\UseRawInputEncoding
\documentclass[runningheads]{llncs}
\usepackage[T1]{fontenc}
\usepackage{graphicx}
\usepackage{adjustbox}
\usepackage{multirow}
\usepackage{amsmath}
\usepackage{amssymb}
\usepackage{booktabs}
\usepackage{array, caption, tabularx,  ragged2e,  booktabs}
\usepackage{comment}

\usepackage{amsfonts}
\usepackage{xcolor}

\usepackage[pagebackref,breaklinks,colorlinks]{hyperref}

\usepackage[capitalize]{cleveref}

\usepackage{todonotes}

\begin{document}
\title{Adjustable Visual Appearance for Generalizable Novel View Synthesis}

\author{Josef Bengtson\inst{1}\and
David Nilsson\inst{1} \and
Che-Tsung Lin\inst{1} \and Marcel B\"usching \inst{2} \and Fredrik Kahl \inst{1}}

\authorrunning{Josef Bengtson et al.}

\institute{Computer Vision Group, Chalmers
University of Technology\\
\email{\{bjosef,david.nilsson,chetsung,fredrik.kahl\}@chalmers.com} \and
KTH Royal Institute of Technology\\
\email{busching@kth.se}}
\maketitle              % typeset the header of the contribution
\begin{abstract}
We present a generalizable novel view synthesis method which enables modifying the visual appearance of an observed scene so rendered views match a target weather or lighting condition without any scene specific training or access to reference views at the target condition. Our method is based on a pretrained generalizable transformer architecture and is fine-tuned on synthetically generated scenes under different appearance conditions. This allows for rendering novel views in a consistent manner for 3D scenes that were not included in the training set, along with the ability to (i) modify their appearance to match the target condition and (ii) smoothly interpolate between different conditions. Experiments on real and synthetic scenes show that our method is able to generate 3D consistent renderings while making realistic appearance changes, including qualitative and quantitative comparisons. Please refer to our project page for video results: \href{https://ava-nvs.github.io}{https://ava-nvs.github.io}

\keywords{3D Style transfer  \and Generalizable Novel View Synthesis  \and NeRFs.}
\end{abstract}
\section{Introduction}
\begin{figure}[tb]
\centering
\includegraphics[width=\columnwidth]{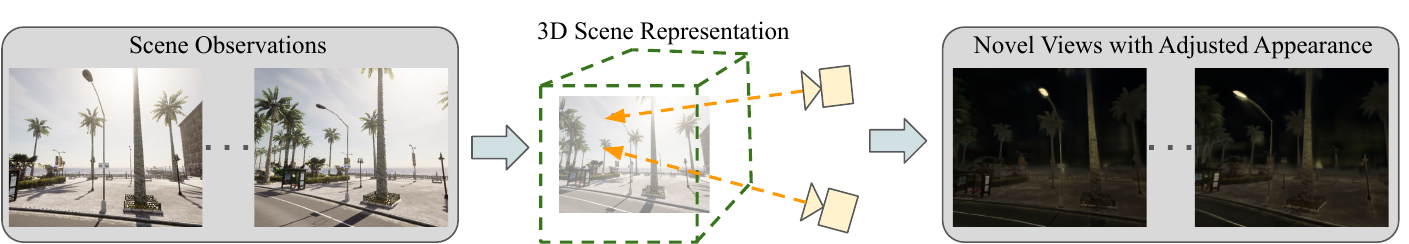}
\caption{Given multiple views of a scene in one weather and lighting condition, we want to generate novel views of the given scene with adjusted visual appearance corresponding to a target condition without scene specific optimization.}
\label{fig_Intro}
\end{figure}

The field of novel view synthesis has seen rapid progress in the last few years after the success of Neural Radiance Fields (NeRFs) \cite{NeRF} and follow-up works \cite{InstantNeuralGraphics,NeRF-W,Mip-NeRF}. A desired quality for these types of 3D scene representations is to be able to disentangle different scene properties from each other, for instance, being able to change the visual appearance without changing the content of the scene. There exists some works in this direction \cite{NeRF-W,Block-NeRF}, but they are limited to interpolating between {\em observed} visual appearances of the 3D scene, thus requiring images of the scene with the desired visual appearance. In contrast, we develop a method that is able to generalize to 3D scenes not used in training, and that thus can adjust the appearance of a scene without having access to any images of that scene at the target visual appearance, see Fig.~\ref{fig_Intro}.

For traditional NeRF-based methods, the properties of the 3D scene are encoded in the weights of a multilayer perceptron (MLP), so each trained model is exclusive to that particular scene. A main challenge is thus that a separate optimization process has to be performed for each individual scene. One approach to handle this is to find ways to improve the efficiency of the training process \cite{InstantNeuralGraphics,AdaNeRF}. A different approach is to avoid per-scene training and instead train cross-scene generalizable methods \cite{pixelNeRF,IBRNet,MVSNeRF,Neuray,GNT}, which are able to synthesize novel views of a scene given just images and corresponding camera poses, and do not require expensive scene-specific optimization.

We present a generalizable novel view synthesis method that allows for changing the visual appearance of a scene while ensuring multi-view consistency. For this we build upon Generalizable NeRF Transformer (GNT) \cite{GNT}, a transformer-based \cite{transformer} novel view synthesis method. Specifically, we introduce a latent appearance variable to enable the control of the visual appearance of rendered views. By using a generalizable NeRF model and the introduced latent appearance variable, we are able to render novel views and change the appearance of scenes that are not seen when training our model without the need for observations of the scene at the target appearance. We will release code and trained models.

In summary, our main contributions are:
\begin{itemize}
	\item We introduce a method that allows for changing the appearance of a novel scene, while ensuring multi-view consistency, by using a latent appearance variable conditioned on a target visual appearance.
	\item We propose a novel loss function which is designed to align the views rendered with a target appearance to the scene observed in that target condition, which enables jointly learning novel view synthesis and appearance change.
	\item We create a synthetic dataset containing urban scenes, with each scene available at four different diverse weather and lighting conditions. The dataset is used for training our model for visual appearance change and enables quantitative evaluation. The dataset will be made publicly available.
\end{itemize}

\section{Related work}
%The recent developments in novel view synthesis originate in the combination of multi-view geometry techniques and deep learning based approaches.
Here we will review progress on NeRFs, focusing on generalizable methods. We will then review 2D style transfer methods and stylized NeRFs methods.
%, finishing with a review of work on changing appearance of and .

%\subsection{Neural Radiance Fields}
\paragraph{Neural Radiance Fields.\ }
NeRFs \cite{NeRF} synthesize consistent and photo-realistic novel views of a scene, by representing each scene as a continuous 5D radiance field parameterized by an MLP mapping 3D positions and 2D viewing directions to volume densities and view-dependent emitted radiances. Views are synthesized by querying points along camera rays and and utilizing volumetric rendering to aggregate the output colors and densities into RGB values. There have been several works improving NeRFs further, e.g. to improve the efficiency \cite{InstantNeuralGraphics,AdaNeRF} and handling few input views \cite{pixelNeRF,RegNeRF}.

\paragraph{Generalizable Novel View Synthesis.\ }
The original NeRF methodology is constrained to training a neural network for representing a single scene, optimizing from scratch for each new scene, without leveraging any prior knowledge. Methods for generalizable neural rendering address this limitation by training on multiple scenes, enabling the learning of a general understanding of how to utilize source observations to synthesize novel views.
Earlier methods such as \cite{pixelNeRF,MVSNeRF} use a multilayer perceptron (MLP) conditioned on feature vectors extracted from the source images to predict color and radiance values which are aggregated with volumetric rendering. To enhance generalization capabilities and rendering quality, recent approaches have incorporated transformer-based architectures \cite{transformer,VIT} for feature aggregation from the source images \cite{Neuray,TransNeRF}, computing densities along the camera ray \cite{IBRNet}, and even for the entire rendering pipeline \cite{GPNR,GNT,SRN,du2023LearningRenderNovel}. While these methods have demonstrated impressive rendering quality, they are currently incapable of modifying the appearance of the rendered views.

\paragraph{2D Style Transfer.\ }

% Alex
The success of Generative Adversarial Networks (GANs)~\cite{goodfellow2014generative} has largely driven advances in 2D style transfer. Methods such as Pix2Pix~\cite{pix2pix2017}, Pix2Pix-HD~\cite{wang2018high}, and BicycleGAN~\cite{zhu2017toward} utilize paired training data, which consists of corresponding images in the source and target conditions. CycleGAN~\cite{CycleGAN2017} and CyEDA~\cite{beh2022cyeda} employ cycle-consistency constraints to learn from unpaired data. NICE-GAN~\cite{chen2020reusing} reuses the discriminator for encoding the images of the target domain. In addition to GANs, the style-attentional network (SANet)~\cite{park2019arbitrary} can synthesize a content image with the style of another image. Diffusion models~\cite{dhariwal2021diffusion,ho2020denoising} have recently achieved superior results in image generation. Palette~\cite{saharia2022palette} introduced the first diffusion-based paired image-translation model, and DiffuseIT~\cite{kwon2022diffusion} recently presented a diffusion-based unsupervised image translation method. Instruct-Pix2Pix~\cite{brooks2022instructpix2pix} re-trains a latent diffusion model~\cite{rombach2022high} using paired images generated by prompt-to-prompt~\cite{hertz2022prompt} and massive instructions generated by GPT-3~\cite{brown2020language} to facilitate instruction-based style transfer. While the images translated by these 2D style transfer methods can individually appear realistic, they do not ensure temporal consistency. In contrast, our method inherently ensures 3D consistency. We experimentally compare our results with 2D style transfer methods applied frame by frame on rendered views.

\paragraph{Visual Appearance Change for NeRF Models. }
Prior work to enable changing the visual appearance of a NeRF model  \cite{NeRF-W,Block-NeRF} typically assign an appearance embedding vector to each image which affect the appearance but not geometry, and is optimized alongside the NeRF model parameters. In \cite{NeRF-W}, low dimensional embeddings allows for smooth interpolation between lighting conditions. One limitation of this approach is that it requires access to images of the scene at both lighting conditions as input. In contrast, our method is a generalizable method that does not require images at both lighting conditions as input when rendering novel views with changed visual appearance.
% It is also limited in the appearance changes it can handle, mainly handling changes in lighting conditions and being unable to adapt the geometry of the scene.
% \paragraph{NeRF Editing.}
% Describe existing work on stylizing NeRFs. Limited to single scene and to non-realistic appearance changes
Another line of research is to edit the style of a NeRF model based on a given style prompt \cite{ArbitraryStyleTransfer,styleNeRF,StylizedNeRF,ARF} typically given as a reference image. More recent works \cite{CLIP-NeRF,NeRF-Art} use the joint language-image embedding space of CLIP \cite{CLIP} to enable specifying the desired style using a text prompt. These methods focus on artistic style changes and have thus not been specifically trained and evaluated on realistic appearance changes such as differences in weather or lighting, in contrast to our method. 
The recent method Instruct-NeRF2NeRF \cite{in2n} enables editing a NeRF model based on a text-prompt, by iteratively updating dataset images using a pretrained 2D editing model \cite{brooks2022instructpix2pix}. This method allows for a variety of appearance changes since they utilize pre-trained diffusion models to perform the editing, but they require training per-scene NeRF models and additional per-scene optimization, in contrast to our method that does not require per-scene training.

\section{Method}
We now give an overview of Generalizable NeRF Transformer (GNT) \cite{GNT} and present our method for adjusting the visual appearance of synthesized views.

\subsection{Basics of GNT}
GNT utilizes a two-stage transformer-based architecture that allows for novel view synthesis from source views. The first stage is a \textit{view transformer} that aggregates information from neighboring views using epipolar geometry. The second stage is a \textit{ray transformer} that performs rendering.

\paragraph{View Transformer. } The \textit{view transformer} computes a coordinate aligned feature field $\mathcal{F} : (\mathbf{x},\mathbf{\theta}) \rightarrow \mathbf{f} \in \mathbb{R}^d $ that maps a 3D position $\mathbf{x}$ and viewing direction $\mathbf{\theta}$ to a feature vector $\mathbf{f}$. Firstly each source view is encoded to a feature map using a U-Net \cite{U-Net} Image Encoder $\mathbf{F}_i = \text{U-Net}(\mathbf{I}_i)$.
The feature representation of a 3D point $\mathbf{x}$ is obtained by projecting it to every source image via the projections $\Pi_i(\mathbf{x})$ and fetching the corresponding value of $\mathbf{F}_i$.
%To obtain the feature representation corresponding to a 3D point $\mathbf{x}$, the point is projected to every source image via the camera projections $\Pi_i(\mathbf{x})$, the corresponding feature values are then calculated by interpolating the image-aligned feature map at the projected point. 
The \textit{view transformer} (VT) is then used to combine all these feature vectors through attention as
\begin{equation}
	\mathcal{F}(\mathbf{x},\mathbf{\theta}) = \text{VT}(\mathbf{F}_1(\Pi_1(\mathbf{x}),\mathbf{\theta}), \cdots, \mathbf{F}_N(\Pi_N(\mathbf{x}),\mathbf{\theta})).
\end{equation}

\paragraph{Ray Transformer.} The \textit{ray transformer} aggregates information along a given camera ray by performing attention between feature values $\mathbf{f}_i = \mathcal{F}(\mathbf{x}_i,\mathbf{\theta})$ on the ray. The GNT pipeline consists of stacking several view and ray transformers, which iteratively refines the feature field. The final \textit{ray transformer}  computes the RGB value $\mathbf{C(r)}$ corresponding to a camera ray $\mathbf{r}$ by feeding the sequence of feature vectors along the ray $\{ \mathbf{f}_1, \cdots, \mathbf{f}_M\}$ into the \textit{ray transformer}, performing mean pooling followed by an MLP as
\begin{equation}
	\mathbf{C(r)} = \text{MLP} \circ \text{Mean} \circ \text{RT}(\mathbf{f}_1, \ldots, \mathbf{f}_M).
\end{equation}

This enables training the method using the standard color prediction loss term commonly used by NeRFs.
%This enables training the method using the standard color prediction loss term, % used in NeRFs,
%\begin{equation}
%	\mathcal{L}_{color} = \left \| \mathbf{C}(r) -  \mathbf{\hat{C}}(r) \right \|_2^2,
%\end{equation}
%where $\mathbf{\hat{C}}(r)$ is the ground-truth color for the pixel corresponding to the ray $\mathbf{r}$.
The attention values from the ray transformer correspond to the importance of each feature vector $\mathbf{f}_i$ along the ray, filling a similar role as the opacity in a traditional NeRF method.
\begin{figure*}[!tb]
\centering
\includegraphics[width=1\textwidth]{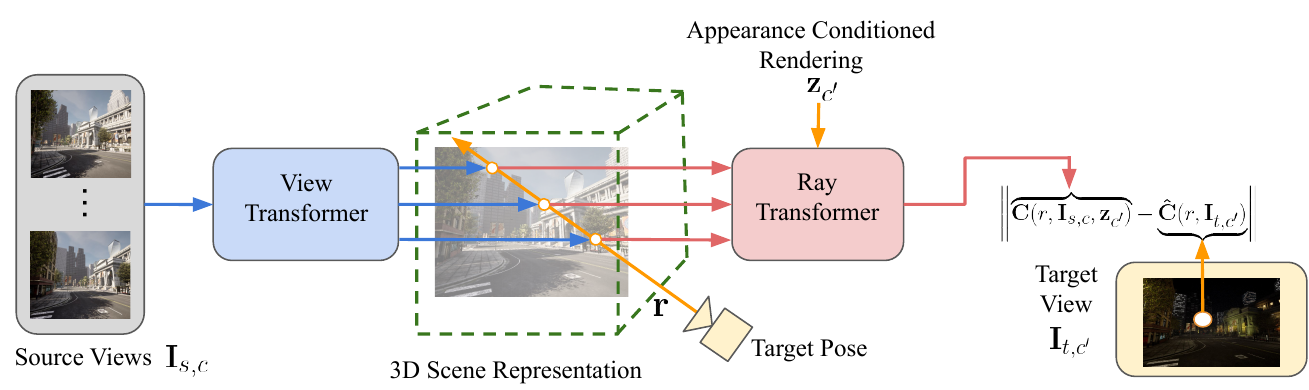}
\caption{Overview of our method for changing visual appearance of synthesized novel views. A target view direction is chosen and camera rays $\mathbf{r}$ are cast and the corresponding source views $\mathbf{I}_{s,c}$ are used to generate a scene representation. A latent appearance variable $\mathbf{z_{c'}}$ is included with the goal of adapting the appearance of the rendered image to match the target view. If the target view is at a different weather or daylight conditions ($c\not=c'$) then this means adapting the visual appearance to match that found in the target view $\mathbf{I}_{t,c'}$ instead of the visual appearance of the source views $\mathbf{I}_{s,c}$.}
\label{fig:Overview}
\end{figure*}

\subsection{Adjusting Visual Appearance}
To change the visual appearance of rendered views to match a target appearance, we propose to introduce a latent appearance variable $\mathbf{z_{c'}}$ as an additional input to the \textit{ray transformer}, to condition the rendering on the target appearance. The proposed architecture can be seen in Fig.~\ref{fig:Overview}.

 The latent variable should correspond to a predefined appearance condition and the value for each condition is jointly optimized with the rest of the network. Since our goal is to change the visual appearance, we include $\mathbf{z_{c'}}$ so that the geometry is kept unchanged. To ensure this, it is used to update the value-tokens in the \textit{ray transformer} while keeping the attention values unchanged, i.e., $V_{c'} = f_z([V;\bf{z}_{c'}])$
% \begin{equation}
% V_{c'} = f_z\begin{pmatrix}
% \begin{bmatrix}
% V\\
% \mathbf{z_{c'}}
% \end{bmatrix}
% \end{pmatrix},
% \end{equation}
where $f_z$ is a single layer MLP that takes in the original value tokens $V$ concatenated with the latent appearance variable $\mathbf{z_{c'}}$ and generates new value tokens $V_{c'}$.
% \begin{equation}
% V = f_v\begin{pmatrix}
% \begin{bmatrix}
% \mathcal{F}(\mathbf{o}+t_\mathbf{d},\mathbf{\theta})\\
% \mathbf{z_{c'}}
% \end{bmatrix}
% \end{pmatrix}.
% \end{equation}
This enables computing a visual appearance change loss,
\begin{equation}
	\mathcal{L}_{appearance} =  \left \| \mathbf{C}(r,\mathbf{I}_{s,c},\mathbf{z}_{c'}) -  \mathbf{\hat{C}}(r, \mathbf{I}_{t,c'}) \right \|_2^2.
	\label{eq:AppearanceLoss}
\end{equation}
This loss enforces that when inputting source views $\mathbf{I}_{s,c}$ from the condition $c$ together with the latent appearance variable $\mathbf{z}_{c'}$ of the condition $c'$, then the predicted color $\mathbf{C}(r,\mathbf{I}_{s,c},\mathbf{z}_{c'})$ should match the ground truth color $\mathbf{\hat{C}}(r, \mathbf{I}_{t,c'})$ for the corresponding target images $\mathbf{I}_{t,c'}$, making it possible for the method to learn to adapt the appearance to match a target condition. If the condition for the target image $\mathbf{I}_{t,c}$ corresponds to that of the source images $\mathbf{I}_{s,c}$, then this becomes a traditional reconstruction loss $ \mathcal{L}_{rec}$, and our full loss is $\mathcal{L}= \mathcal{L}_{rec} + \mathcal{L}_{appearance}$. 
% \begin{equation}
% 	\mathcal{L}_{rec} =  \left \| \mathbf{C}(r,\mathbf{I}_{s,c},\mathbf{z}_c) -  \mathbf{\hat{C}}(r, \mathbf{I}_{t,c}) \right \|_2^2.
% 	\label{eq:Lrec}
% \end{equation}
% The final loss term is then acquired by combining these two loss terms,
% \begin{equation}
% 	\mathcal{L}= \mathcal{L}_{rec} + \mathcal{L}_{appearance}.
% 	\label{eq:FullLoss}
% \end{equation}
Rendering images with changed visual appearance is done by computing the rendered color $\mathbf{C}(r,I_{s,c},\mathbf{z}_{c'})$ for all pixels in an image, giving as input source views from one condition and a latent variable $\mathbf{z}_{c'}$ corresponding to the desired target condition.

\begin{figure}[t]
\centering
\includegraphics[width=1\columnwidth]{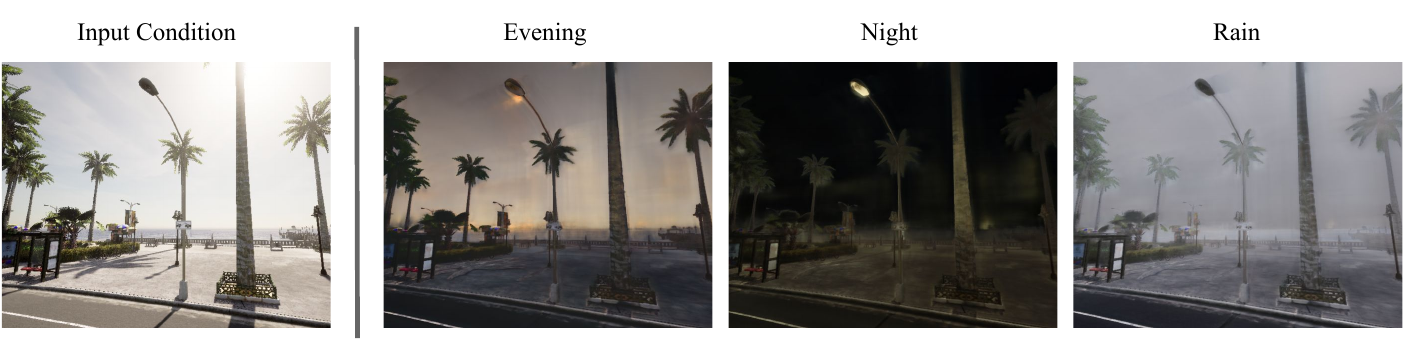}
\caption{Appearance change from the day condition into three other conditions. We observe that our method is able to take images at one condition and generate new views of that scene at the three other conditions by changing the overall visual appearance of the images to match the desired condition and by making local changes such as turning on street lamps.}
\label{fig_results_4seasons}
\end{figure}
\begin{figure*}[!t]
\centering
\includegraphics[width=\textwidth]{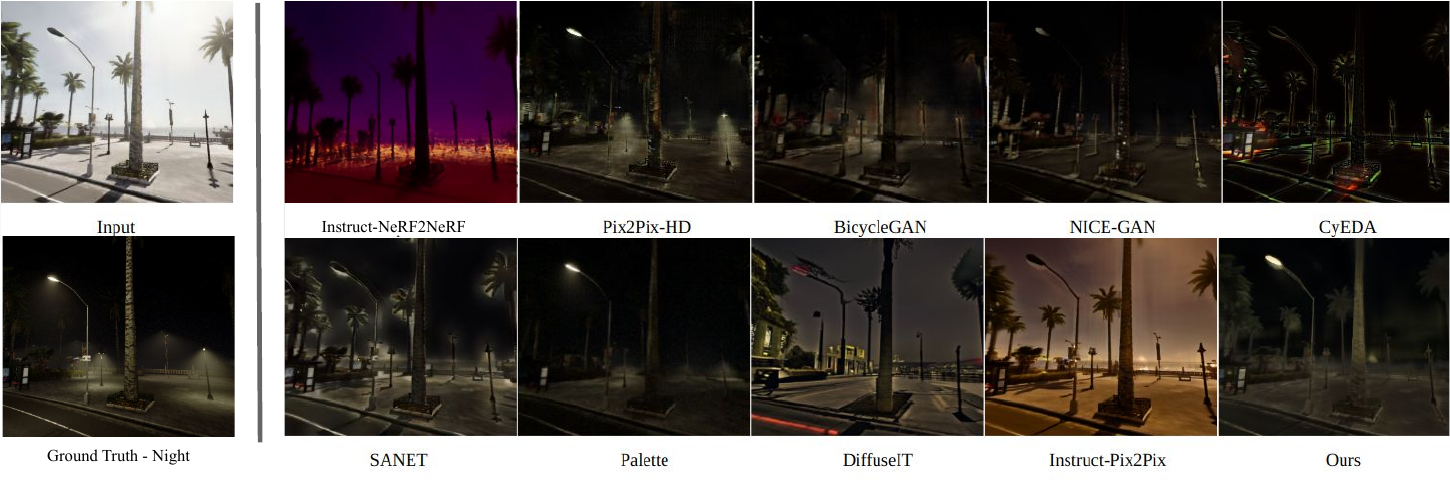}
\caption{Comparing our method with Instruct-NeRF2NeRF \cite{in2n} as well as applying different 2D style transfer methods on rendered images.
We note that Instruct-Pix2Pix~\cite{brooks2022instructpix2pix} effectively generates realistic 2D edits; however, it exhibits significant inconsistencies that Instruct-NeRF2NeRF fails to
consolidate in 3D, leading to an unrealistic appearance.
Only our method, Pix2Pix-HD~\cite{wang2018high} and Palette~\cite{saharia2022palette} learn to turn on the street lamps. SANet~\cite{park2019arbitrary} and CyEDA~\cite{beh2022cyeda} achieve better structure preservation with some noticeable artifacts. The diffusion models DiffuseIT~\cite{kwon2022diffusion} and Instruct-Pix2Pix~\cite{brooks2022instructpix2pix} can provide visually plausible results for individual images, but there are hallucinations that do not exist in the original images, leading to multi-view inconsistencies. Palette provides more realistic images, but it is however lacking in temporal consistency. Comparisons for additional conditions can be found in appendix \ref{sec:MoreScenarios}.}
\label{fig_2dcomp}
\end{figure*}
\section{Experiments}
Qualitative and quantitative experiments are performed to test our method's ability to adapt the visual appearance of real and synthetic scenes that have not been seen during training.
\paragraph{Dataset.} The used dataset is generated using the autonomous driving simulator CARLA \cite{CARLA}, which enables the generation of synthetic images within a simulated city environment along with their ground truth camera poses. Additionally, weather and lighting conditions can easily be changed. 

For our experiments, four conditions were defined, corresponding to \textit{night}, \textit{day}, \textit{rain} and \textit{evening}. A scene was defined as a sequence of 10 observations taken along a road. With four different conditions, this led to a total of 40 images per scene. All generated images are $800 \times 600$ pixels. The CARLA map was split into two regions, one to generate 145 training scenes, and the other to generate 38 evaluation scenes, ensuring separation between training and evaluation scenes. This dataset alongside the code will be made publicly available. We also show qualitative examples evaluating our trained model on scenes from the Spaces dataset~\cite{Spaces} to show that our method can generalize to real images.

% , with small pose differences between images
\paragraph{Implementation.}
The model was initialized with weights from a GNT network pretrained on a combination of synthetic and real data \cite{GNT}. The model was trained to perform visual appearance change using the 145 training scenes from the introduced CARLA dataset, including the proposed appearance change loss term \eqref{eq:AppearanceLoss}. The training was performed on a single A100 GPU, taking approximately 8 hours, and the method was then able to generalize to scenes not seen during training. Note that the model was trained for all training scenes at the same time, and there is no scene-specific training for the test scenes. When we test the model, we only use images of the test scene in the source condition $\mathbf{I}_{s,c}$ and not any images of the test scene in the target condition $c' \neq c$.

\paragraph{Baselines.}
The GAN-based methods, such as Pix2Pix-HD~\cite{wang2018high}, BicycleGAN~\cite{zhu2017toward}, NICE-GAN\cite{chen2020reusing}, and CyEDA~\cite{beh2022cyeda}, along with the diffusion model-based method Palette~\cite{saharia2022palette}, have been retrained using our synthetic dataset. The reference-based methods, DiffuseIT~\cite{kwon2022diffusion} and SANet~\cite{park2019arbitrary}, are capable of performing image translation using a reference images at the target condition from the synthetic dataset. Instruct-Pix2Pix \cite{brooks2022instructpix2pix} is pre-trained on editing images based on text prompts and was not retrained on our synthetic dataset. 

Furthermore, we compared our method with the Instruct-NeRF2NeRF \cite{in2n} model, utilizing the official implementation that employs the Nerfstudio \cite{Nerfstudio} Nerfacto NeRF model. Due to the unsatisfactory quality of the Nerfacto models when using 10 images, we increased the number of images in the sequence to 25 images. More details can be found in appendix \ref{sec:in2n_comp}.

\setlength{\tabcolsep}{5pt}
\begin{table}[t]
\caption{Comparison of similarity of rendered views for our method with ground truth images for all combinations of weather and lighting conditions (\textbf{\small{PSNR$\uparrow$ $\vert$ SSIM$\uparrow$ $\vert$ LPIPS$\downarrow$}}). The values along the diagonal correspond to novel view synthesis without appearance change. The off-diagonal values correspond to evaluating novel views with changed visual appearance to match the target condition.}
\centering
%\small
\begin{adjustbox}{width=\linewidth,center}
\begin{tabular}{lcccc}
\cmidrule[2pt]{1-5}
%\hline
         	& From Day                                                                                             	& From Night                                                                                           	& From Evening                                                                                         	& From Rain                                                                                            	\\ \cmidrule[2pt]{1-5} %\hline
Into Day 	& 23.9 \hspace{-0.55em} \textcolor{gray}{$\vert$}\hspace{-0.25em} 0.77  \hspace{-0.55em} \textcolor{gray}{$\vert$}\hspace{-0.25em} 0.60  & 15.3  \hspace{-0.55em} \textcolor{gray}{$\vert$}\hspace{-0.25em} 0.56  \hspace{-0.55em} \textcolor{gray}{$\vert$}\hspace{-0.25em} 0.62 & 16.7  \hspace{-0.55em} \textcolor{gray}{$\vert$}\hspace{-0.25em} 0.64  \hspace{-0.55em} \textcolor{gray}{$\vert$}\hspace{-0.25em} 0.61 & 15.7  \hspace{-0.55em} \textcolor{gray}{$\vert$}\hspace{-0.25em} 0.59  \hspace{-0.55em} \textcolor{gray}{$\vert$}\hspace{-0.25em} 0.60 \\
Into Night   & 21.0  \hspace{-0.55em} \textcolor{gray}{$\vert$}\hspace{-0.25em} 0.56  \hspace{-0.55em} \textcolor{gray}{$\vert$}\hspace{-0.25em} 0.55 & 27.4  \hspace{-0.55em} \textcolor{gray}{$\vert$}\hspace{-0.25em} 0.68  \hspace{-0.55em} \textcolor{gray}{$\vert$}\hspace{-0.25em} 0.57 & 20.7  \hspace{-0.55em} \textcolor{gray}{$\vert$}\hspace{-0.25em} 0.54  \hspace{-0.55em} \textcolor{gray}{$\vert$}\hspace{-0.25em} 0.55 & 21.2  \hspace{-0.55em} \textcolor{gray}{$\vert$}\hspace{-0.25em} 0.57  \hspace{-0.55em} \textcolor{gray}{$\vert$}\hspace{-0.25em} 0.55 \\
Into Evening & 24.1  \hspace{-0.55em} \textcolor{gray}{$\vert$}\hspace{-0.25em} 0.75  \hspace{-0.55em} \textcolor{gray}{$\vert$}\hspace{-0.25em} 0.58 & 20.0  \hspace{-0.55em} \textcolor{gray}{$\vert$}\hspace{-0.25em} 0.62  \hspace{-0.55em} \textcolor{gray}{$\vert$}\hspace{-0.25em} 0.57 & 25.4  \hspace{-0.55em} \textcolor{gray}{$\vert$}\hspace{-0.25em} 0.76  \hspace{-0.55em} \textcolor{gray}{$\vert$}\hspace{-0.25em} 0.58 & 21.4  \hspace{-0.55em} \textcolor{gray}{$\vert$}\hspace{-0.25em} 0.69  \hspace{-0.55em} \textcolor{gray}{$\vert$}\hspace{-0.25em} 0.57 \\
Into Rain	& 23.4  \hspace{-0.55em} \textcolor{gray}{$\vert$}\hspace{-0.25em} 0.71  \hspace{-0.55em} \textcolor{gray}{$\vert$}\hspace{-0.25em} 0.58 & 21.7  \hspace{-0.55em} \textcolor{gray}{$\vert$}\hspace{-0.25em} 0.66  \hspace{-0.55em} \textcolor{gray}{$\vert$}\hspace{-0.25em} 0.57 & 21.3  \hspace{-0.55em} \textcolor{gray}{$\vert$}\hspace{-0.25em} 0.69  \hspace{-0.55em} \textcolor{gray}{$\vert$}\hspace{-0.25em} 0.56 & 26.8  \hspace{-0.55em} \textcolor{gray}{$\vert$}\hspace{-0.25em} 0.78  \hspace{-0.55em} \textcolor{gray}{$\vert$}\hspace{-0.25em} 0.58 \\ \cmidrule[2pt]{1-5} %\hline
\end{tabular}
\end{adjustbox}
\label{tab:results_2D_comp}
\end{table}
\setlength{\tabcolsep}{3pt}

\begin{table*}[t]
\caption{Qualitative comparison of rendering quality against 2D style transfer methods (\textbf{\small{PSNR$\uparrow$ $\vert$ SSIM$\uparrow$ $\vert$ LPIPS$\downarrow$}}). We observe that our method outperforms all 2D style transfer methods on these metrics, with significant increases in performance on PSNR and SSIM for most scenarios}
\begin{minipage}{\textwidth}
\begin{adjustbox}{width=\linewidth,center}
\begin{tabular}{clcccccc}
%\hline
\cmidrule[3pt]{1-8}
\multirow{2}{*}{Type}                                                                         	& \multicolumn{1}{c}{\multirow{2}{*}{Method}} & \multicolumn{6}{c}{Scenarios}                                                                                                                         	\\ \cline{3-8}
                                                                                              	& \multicolumn{1}{c}{}                    	& \multicolumn{1}{c}{Day to Night} & \multicolumn{1}{c}{Day to Evening} & \multicolumn{1}{c}{Day to Rain} & \multicolumn{1}{c}{Night to Day} & \multicolumn{1}{c}{Evening to Day} & \multicolumn{1}{c}{Rain to Day} \\ \hline
\multicolumn{1}{c|}{\multirow{5}{*}{\begin{tabular}[c]{@{}c@{}}Non- \\ diffusion\end{tabular}}}                                                    	& Pix2Pix-HD~\cite{wang2018high}                              	&   19.7 \hspace{-0.55em} \textcolor{gray}{$\vert$}\hspace{-0.25em} 0.36 \hspace{-0.55em} \textcolor{gray}{$\vert$}\hspace{-0.25em} 0.565 & 18.4  \hspace{-0.55em} \textcolor{gray}{$\vert$}\hspace{-0.25em} 0.35  \hspace{-0.55em} \textcolor{gray}{$\vert$}\hspace{-0.25em} 0.603 & 19.7  \hspace{-0.55em} \textcolor{gray}{$\vert$}\hspace{-0.25em} 0.53  \hspace{-0.55em} \textcolor{gray}{$\vert$}\hspace{-0.25em} 0.582 & 13.8  \hspace{-0.55em} \textcolor{gray}{$\vert$}\hspace{-0.25em} 0.40  \hspace{-0.55em} \textcolor{gray}{$\vert$}\hspace{-0.25em} 0.629 & 15.3  \hspace{-0.55em} \textcolor{gray}{$\vert$}\hspace{-0.25em} 0.46  \hspace{-0.55em} \textcolor{gray}{$\vert$}\hspace{-0.25em} 0.619 & 13.9  \hspace{-0.55em} \textcolor{gray}{$\vert$}\hspace{-0.25em} 0.43  \hspace{-0.55em} \textcolor{gray}{$\vert$}\hspace{-0.25em} 0.629 \\
\multicolumn{1}{c|}{}                                                              	 
    	& BicycleGAN~\cite{zhu2017toward}                              	& 19.0  \hspace{-0.55em} \textcolor{gray}{$\vert$}\hspace{-0.25em} 0.38  \hspace{-0.55em} \textcolor{gray}{$\vert$}\hspace{-0.25em} 0.556 & 18.8  \hspace{-0.55em} \textcolor{gray}{$\vert$}\hspace{-0.25em} 0.41  \hspace{-0.55em} \textcolor{gray}{$\vert$}\hspace{-0.25em} 0.587 & 22.7  \hspace{-0.55em} \textcolor{gray}{$\vert$}\hspace{-0.25em} 0.66  \hspace{-0.55em} \textcolor{gray}{$\vert$}\hspace{-0.25em} 0.578 & 14.2  \hspace{-0.55em} \textcolor{gray}{$\vert$}\hspace{-0.25em} 0.47  \hspace{-0.55em} \textcolor{gray}{$\vert$}\hspace{-0.25em} 0.630 & 15.9  \hspace{-0.55em} \textcolor{gray}{$\vert$}\hspace{-0.25em} 0.56  \hspace{-0.55em} \textcolor{gray}{$\vert$}\hspace{-0.25em} 0.627 & 15.0  \hspace{-0.55em} \textcolor{gray}{$\vert$}\hspace{-0.25em} 0.54  \hspace{-0.55em} \textcolor{gray}{$\vert$}\hspace{-0.25em} 0.630 \\
\multicolumn{1}{c|}{}                                                              	 
    	& NICE-GAN~\cite{chen2020reusing}                                	& 18.3  \hspace{-0.55em} \textcolor{gray}{$\vert$}\hspace{-0.25em} 0.29  \hspace{-0.55em} \textcolor{gray}{$\vert$}\hspace{-0.25em} 0.553 & 18.8  \hspace{-0.55em} \textcolor{gray}{$\vert$}\hspace{-0.25em} 0.39  \hspace{-0.55em} \textcolor{gray}{$\vert$}\hspace{-0.25em} 0.589 & 20.8  \hspace{-0.55em} \textcolor{gray}{$\vert$}\hspace{-0.25em} 0.56  \hspace{-0.55em} \textcolor{gray}{$\vert$}\hspace{-0.25em} 0.583 & 12.9  \hspace{-0.55em} \textcolor{gray}{$\vert$}\hspace{-0.25em} 0.29  \hspace{-0.55em} \textcolor{gray}{$\vert$}\hspace{-0.25em} 0.626 & 14.6  \hspace{-0.55em} \textcolor{gray}{$\vert$}\hspace{-0.25em} 0.45  \hspace{-0.55em} \textcolor{gray}{$\vert$}\hspace{-0.25em} 0.618 & 14.3  \hspace{-0.55em} \textcolor{gray}{$\vert$}\hspace{-0.25em} 0.47  \hspace{-0.55em} \textcolor{gray}{$\vert$}\hspace{-0.25em} 0.624 \\
\multicolumn{1}{c|}{}                                                                         	& CyEDA~\cite{beh2022cyeda}                                   	& 17.9  \hspace{-0.55em} \textcolor{gray}{$\vert$}\hspace{-0.25em} 0.32  \hspace{-0.55em} \textcolor{gray}{$\vert$}\hspace{-0.25em} 0.556 & 18.8  \hspace{-0.55em} \textcolor{gray}{$\vert$}\hspace{-0.25em} 0.40  \hspace{-0.55em} \textcolor{gray}{$\vert$}\hspace{-0.25em} 0.597 & 20.0  \hspace{-0.55em} \textcolor{gray}{$\vert$}\hspace{-0.25em} 0.67  \hspace{-0.55em} \textcolor{gray}{$\vert$}\hspace{-0.25em} 0.579 & 11.7  \hspace{-0.55em} \textcolor{gray}{$\vert$}\hspace{-0.25em} 0.47  \hspace{-0.55em} \textcolor{gray}{$\vert$}\hspace{-0.25em} 0.625 & 14.0  \hspace{-0.55em} \textcolor{gray}{$\vert$}\hspace{-0.25em} 0.59  \hspace{-0.55em} \textcolor{gray}{$\vert$}\hspace{-0.25em} 0.633 & 12.3  \hspace{-0.55em} \textcolor{gray}{$\vert$}\hspace{-0.25em} 0.53  \hspace{-0.55em} \textcolor{gray}{$\vert$}\hspace{-0.25em} 0.634 \\
\multicolumn{1}{c|}{}                                                                         	& SANet~\cite{park2019arbitrary}                                   	& 18.9  \hspace{-0.55em} \textcolor{gray}{$\vert$}\hspace{-0.25em} 0.50  \hspace{-0.55em} \textcolor{gray}{$\vert$}\hspace{-0.25em} 0.571 & 20.2  \hspace{-0.55em} \textcolor{gray}{$\vert$}\hspace{-0.25em} 0.64  \hspace{-0.55em} \textcolor{gray}{$\vert$}\hspace{-0.25em} 0.606 & 20.1  \hspace{-0.55em} \textcolor{gray}{$\vert$}\hspace{-0.25em} 0.66  \hspace{-0.55em} \textcolor{gray}{$\vert$}\hspace{-0.25em} 0.581 & 14.5  \hspace{-0.55em} \textcolor{gray}{$\vert$}\hspace{-0.25em} 0.52  \hspace{-0.55em} \textcolor{gray}{$\vert$}\hspace{-0.25em} 0.629 & 15.5  \hspace{-0.55em} \textcolor{gray}{$\vert$}\hspace{-0.25em} 0.59  \hspace{-0.55em} \textcolor{gray}{$\vert$}\hspace{-0.25em} 0.618 & 12.6  \hspace{-0.55em} \textcolor{gray}{$\vert$}\hspace{-0.25em} 0.45  \hspace{-0.55em} \textcolor{gray}{$\vert$}\hspace{-0.25em} 0.616 \\ \hline
\multicolumn{1}{c|}{\multirow{3}{*}{\begin{tabular}[c]{@{}c@{}}Diffusion \\ Models\end{tabular}}} & Palette~\cite{saharia2022palette}                                 	& 19.4  \hspace{-0.55em} \textcolor{gray}{$\vert$}\hspace{-0.25em} 0.42  \hspace{-0.55em} \textcolor{gray}{$\vert$}\hspace{-0.25em} 0.577 & 20.6  \hspace{-0.55em} \textcolor{gray}{$\vert$}\hspace{-0.25em} 0.54  \hspace{-0.55em} \textcolor{gray}{$\vert$}\hspace{-0.25em} 0.618 & 22.6  \hspace{-0.55em} \textcolor{gray}{$\vert$}\hspace{-0.25em} 0.66  \hspace{-0.55em} \textcolor{gray}{$\vert$}\hspace{-0.25em} 0.601 & 12.1  \hspace{-0.55em} \textcolor{gray}{$\vert$}\hspace{-0.25em} 0.41  \hspace{-0.55em} \textcolor{gray}{$\vert$}\hspace{-0.25em} 0.689 & \hspace{0.2em} 9.8  \hspace{-0.55em} \textcolor{gray}{$\vert$}\hspace{-0.25em} 0.38  \hspace{-0.55em} \textcolor{gray}{$\vert$}\hspace{-0.25em} 0.688 & \hspace{0.4em}9.8  \hspace{-0.55em} \textcolor{gray}{$\vert$}\hspace{-0.25em} 0.38  \hspace{-0.55em} \textcolor{gray}{$\vert$}\hspace{-0.25em} 0.695 \\
\multicolumn{1}{c|}{}                                                                         	& DiffuseIT~\cite{kwon2022diffusion}                               	& 17.1  \hspace{-0.55em} \textcolor{gray}{$\vert$}\hspace{-0.25em} 0.32  \hspace{-0.55em} \textcolor{gray}{$\vert$}\hspace{-0.25em} 0.594 & 17.2  \hspace{-0.55em} \textcolor{gray}{$\vert$}\hspace{-0.25em} 0.44  \hspace{-0.55em} \textcolor{gray}{$\vert$}\hspace{-0.25em} 0.627 & 16.0  \hspace{-0.55em} \textcolor{gray}{$\vert$}\hspace{-0.25em} 0.43  \hspace{-0.55em} \textcolor{gray}{$\vert$}\hspace{-0.25em} 0.613 & 11.2  \hspace{-0.55em} \textcolor{gray}{$\vert$}\hspace{-0.25em} 0.35  \hspace{-0.55em} \textcolor{gray}{$\vert$}\hspace{-0.25em} 0.626 & 12.3  \hspace{-0.55em} \textcolor{gray}{$\vert$}\hspace{-0.25em} 0.38  \hspace{-0.55em} \textcolor{gray}{$\vert$}\hspace{-0.25em} 0.618 & 11.8  \hspace{-0.55em} \textcolor{gray}{$\vert$}\hspace{-0.25em} 0.35  \hspace{-0.55em} \textcolor{gray}{$\vert$}\hspace{-0.25em} 0.625 \\
\multicolumn{1}{c|}{}                                                                         	& Instruct-Pix2Pix~\cite{brooks2022instructpix2pix}\footnotemark[1]                       	& 15.9  \hspace{-0.55em} \textcolor{gray}{$\vert$}\hspace{-0.25em} 0.34  \hspace{-0.55em} \textcolor{gray}{$\vert$}\hspace{-0.25em} 0.579      	& 14.3  \hspace{-0.55em} \textcolor{gray}{$\vert$}\hspace{-0.25em} 0.53  \hspace{-0.55em} \textcolor{gray}{$\vert$}\hspace{-0.25em} 0.653      	& 14.1  \hspace{-0.55em} \textcolor{gray}{$\vert$}\hspace{-0.25em} 0.53  \hspace{-0.55em} \textcolor{gray}{$\vert$}\hspace{-0.25em} 0.638      	& 11.5  \hspace{-0.55em} \textcolor{gray}{$\vert$}\hspace{-0.25em} 0.46  \hspace{-0.55em} \textcolor{gray}{$\vert$}\hspace{-0.25em} 0.647      	& \hspace{0.2em} 8.7  \hspace{-0.55em} \textcolor{gray}{$\vert$}\hspace{-0.25em} 0.34  \hspace{-0.55em} \textcolor{gray}{$\vert$}\hspace{-0.25em} 0.674      	& 13.4  \hspace{-0.55em} \textcolor{gray}{$\vert$}\hspace{-0.25em} 0.52  \hspace{-0.55em} \textcolor{gray}{$\vert$}\hspace{-0.25em} 0.640      	\\ \hline
\multicolumn{1}{c|}{}                                                                         	& Ours                                    	& \textbf{21.0}  \hspace{-0.55em} \textcolor{gray}{$\vert$}\hspace{-0.25em} \textbf{0.56}  \hspace{-0.55em} \textcolor{gray}{$\vert$}\hspace{-0.25em} \textbf{0.549} & \textbf{24.1}  \hspace{-0.55em} \textcolor{gray}{$\vert$}\hspace{-0.25em} \textbf{0.75}  \hspace{-0.55em} \textcolor{gray}{$\vert$}\hspace{-0.25em} \textbf{0.585} & \textbf{23.4}  \hspace{-0.55em} \textcolor{gray}{$\vert$}\hspace{-0.25em} \textbf{0.71}  \hspace{-0.55em} \textcolor{gray}{$\vert$}\hspace{-0.25em} \textbf{0.577} & \textbf{15.3}  \hspace{-0.55em} \textcolor{gray}{$\vert$}\hspace{-0.25em} \textbf{0.56}  \hspace{-0.55em} \textcolor{gray}{$\vert$}\hspace{-0.25em} \textbf{0.624} & \textbf{16.7}  \hspace{-0.55em} \textcolor{gray}{$\vert$}\hspace{-0.25em} \textbf{0.64}  \hspace{-0.55em} \textcolor{gray}{$\vert$}\hspace{-0.25em} \textbf{0.615} & \textbf{15.7}  \hspace{-0.55em} \textcolor{gray}{$\vert$}\hspace{-0.25em} \textbf{0.59}  \hspace{-0.55em} \textcolor{gray}{$\vert$}\hspace{-0.25em} \textbf{0.602} \\ 
\cmidrule[3pt]{1-8}
%\hline
\end{tabular}
\label{tab:Compare_w_2D}
\end{adjustbox}
\end{minipage}
\end{table*}

\footnotetext[1]{Instruct-Pix2Pix is pre-trained on editing images based on text prompts and was therefore not retrained on our synthetic dataset.}

% Consistency Metric Table With IN2N
\begin{table*}[t]
\caption{Quantitative comparison of the consistency of novel view rendering against 2D style transfer methods and instruct-NeRF2NeRF (\textbf{\small{tOF$\downarrow$ \textcolor{gray}{$\vert$}  tLP$\downarrow$}} \cite{chu2020learning}). We can observe that our method significantly outperforms most of the 2D methods. %, with the most consistent 2D-methods giving similar metrics.
Please see our project page for a video illustrating the rendering consistency: \href{https://ava-nvs.github.io}{https://ava-nvs.github.io}}
%\small
\begin{minipage}{\textwidth}
\centering
\scalebox{0.8}{
% \begin{adjustbox}{width=\linewidth,center}

\begin{tabular}{clccccrrrr}
%\hline
\cmidrule[2pt]{1-6}
\multirow{2}{*}{Type} & \multicolumn{1}{c}{\multirow{2}{*}{Method}} & \multicolumn{4}{c}{Scenarios} \\
\cline{3-6}
& \multicolumn{1}{c}{} & \multicolumn{1}{c}{Day to Night } & \multicolumn{1}{c}{\hspace{0.5em}Day to Evening} & \multicolumn{1}{c}{\hspace{0.5em}Day to Rain} & \multicolumn{1}{c}{\hspace{0.5em}Night to Day} \\
\cmidrule[2pt]{1-6}
\multicolumn{1}{c|}{\multirow{5}{*}{Non-diffusion}} & Pix2Pix-HD~\cite{wang2018high} & 2.59\hspace{-0.32em} \textcolor{gray}{$\vert$}\hspace{-0.27em} 0.147 & 1.58\hspace{-0.32em} \textcolor{gray}{$\vert$}\hspace{-0.27em} 0.169 & 1.60\hspace{-0.32em} \textcolor{gray}{$\vert$}\hspace{-0.27em} 0.030 & 2.49\hspace{-0.32em} \textcolor{gray}{$\vert$}\hspace{-0.27em} 0.078 \\
\multicolumn{1}{c|}{} & BicycleGAN~\cite{zhu2017toward} & 5.13\hspace{-0.32em} \textcolor{gray}{$\vert$}\hspace{-0.27em} 0.053 & 4.79\hspace{-0.32em} \textcolor{gray}{$\vert$}\hspace{-0.27em} 0.083 & 5.10\hspace{-0.32em} \textcolor{gray}{$\vert$}\hspace{-0.27em} 0.024 & 5.20\hspace{-0.32em} \textcolor{gray}{$\vert$}\hspace{-0.27em} 0.047 \\
\multicolumn{1}{c|}{} & NICE-GAN~\cite{chen2020reusing} & 1.93\hspace{-0.32em} \textcolor{gray}{$\vert$}\hspace{-0.27em} 0.040 & 1.24\hspace{-0.32em} \textcolor{gray}{$\vert$}\hspace{-0.27em} 0.081 & 1.25\hspace{-0.32em} \textcolor{gray}{$\vert$}\hspace{-0.27em} 0.014 & 2.09\hspace{-0.32em} \textcolor{gray}{$\vert$}\hspace{-0.27em} 0.055 \\
\multicolumn{1}{c|}{} & CyEDA~\cite{beh2022cyeda} & 1.62\hspace{-0.32em} \textcolor{gray}{$\vert$}\hspace{-0.27em} 0.027 & 1.21\hspace{-0.32em} \textcolor{gray}{$\vert$}\hspace{-0.27em} 0.115 & \textbf{0.96}  \hspace{-0.25em}\textcolor{gray}{$\vert$} 0.022 & \textbf{1.25}\hspace{-0.32em} \textcolor{gray}{$\vert$}\hspace{-0.27em} \textbf{0.032} \\
\multicolumn{1}{c|}{} & SANet~\cite{park2019arbitrary} & 2.37\hspace{-0.32em} \textcolor{gray}{$\vert$}\hspace{-0.27em} 0.069 & 2.05\hspace{-0.32em} \textcolor{gray}{$\vert$}\hspace{-0.27em} 0.097 & 1.73\hspace{-0.32em} \textcolor{gray}{$\vert$}\hspace{-0.27em} 0.088 & 2.01\hspace{-0.32em} \textcolor{gray}{$\vert$}\hspace{-0.27em} 0.092 \\
\cmidrule[2pt]{1-6} %\hline
\multicolumn{1}{c|}{\multirow{3}{*}{\begin{tabular}[c]{@{}c@{}}Diffusion \\ Models\end{tabular}}} & Palette~\cite{saharia2022palette} & \negthickspace \negthickspace 10.12\hspace{-0.32em} \textcolor{gray}{$\vert$}\hspace{-0.27em} 0.115 & 7.21\hspace{-0.32em} \textcolor{gray}{$\vert$}\hspace{-0.27em} 0.109 & 8.41\hspace{-0.32em} \textcolor{gray}{$\vert$}\hspace{-0.27em} 0.057 & \negthickspace \negthickspace 18.77\hspace{-0.32em} \textcolor{gray}{$\vert$}\hspace{-0.27em} 0.050 \\
\multicolumn{1}{c|}{} & DiffuseIT~\cite{kwon2022diffusion} & \negthickspace \negthickspace 24.62\hspace{-0.32em} \textcolor{gray}{$\vert$}\hspace{-0.27em} 0.236 & \negthickspace \negthickspace 29.23\hspace{-0.32em} \textcolor{gray}{$\vert$}\hspace{-0.27em} 0.242 & \negthickspace \negthickspace 27.43\hspace{-0.32em} \textcolor{gray}{$\vert$}\hspace{-0.27em} 0.166 & \negthickspace \negthickspace 29.75\hspace{-0.32em} \textcolor{gray}{$\vert$}\hspace{-0.27em} 0.188 \\
\multicolumn{1}{c|}{} & Instruct-Pix2Pix~\cite{brooks2022instructpix2pix} & 1.62\hspace{-0.32em} \textcolor{gray}{$\vert$}\hspace{-0.27em} 0.123 & 1.37\hspace{-0.32em} \textcolor{gray}{$\vert$}\hspace{-0.27em} 0.121 & 1.45\hspace{-0.32em} \textcolor{gray}{$\vert$}\hspace{-0.27em} 0.088 & 1.48\hspace{-0.32em} \textcolor{gray}{$\vert$}\hspace{-0.27em} 0.092 \\
\cmidrule[2pt]{1-6} %\hline
\multicolumn{1}{c|}{\multirow{1}{*}{NeRF Editing}} & IN2N ~\cite{in2n} & 7.29\hspace{-0.32em} \textcolor{gray}{$\vert$}\hspace{-0.27em} 0.137 & 6.09\hspace{-0.32em} \textcolor{gray}{$\vert$}\hspace{-0.27em} 0.085& 5.65\hspace{-0.32em} \textcolor{gray}{$\vert$}\hspace{-0.27em} 0.051 & \hspace{0.2em}-  \hspace{0.6em}\textcolor{gray}{$\vert$} 
\hspace{0.2em} - \footnotemark[1]\\
\cmidrule[2pt]{1-6} %\hline
\multicolumn{1}{c|}{\multirow{1}{*}{Ours}} & Ours & \textbf{1.44}\hspace{-0.32em} \textcolor{gray}{$\vert$}\hspace{-0.27em} \textbf{0.026} & \textbf{1.10}\hspace{-0.32em} \textcolor{gray}{$\vert$}\hspace{-0.27em} \textbf{0.035} & \hspace{0.3em}0.97\hspace{-0.27em} \textcolor{gray}{$\vert$}\hspace{-0.27em} \textbf{0.013} & \textbf{1.25}\hspace{-0.32em} \textcolor{gray}{$\vert$}\hspace{-0.27em} \textbf{0.032} \\
\cmidrule[2pt]{1-6} %\hline
\end{tabular}
% \end{adjustbox}

}
\end{minipage}

\label{tab:Compare_w_2D_consistency}
\end{table*}

\begin{figure}[t]
\centering
\includegraphics[width=1\columnwidth]{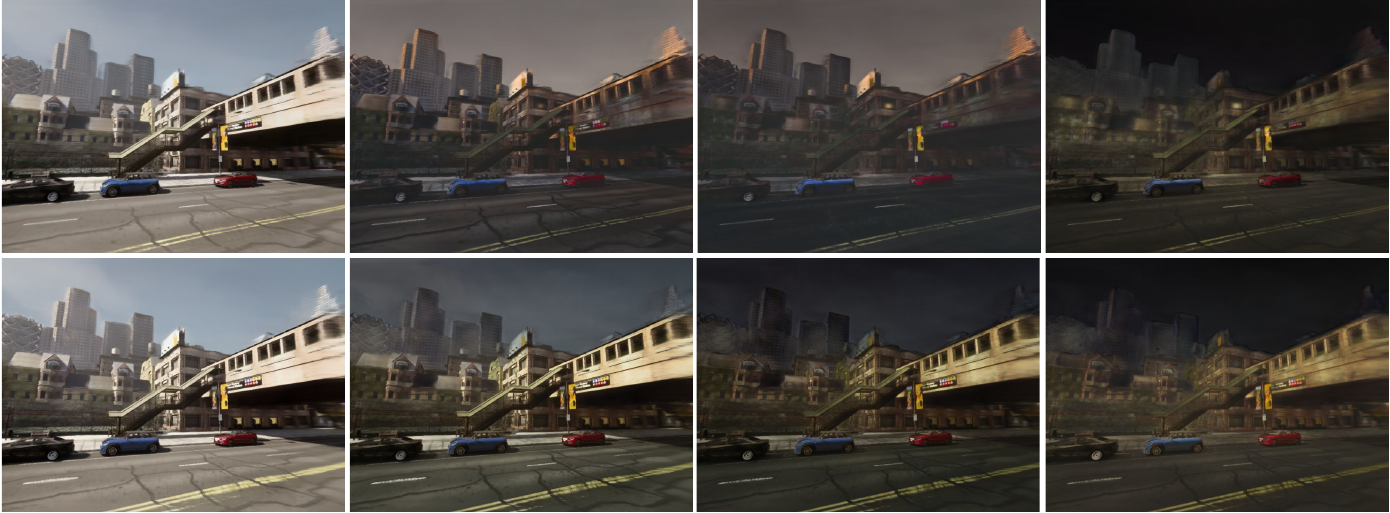}
\caption{Gradually changing visual appearance by interpolating between latent appearance variables corresponding to day and night. The first row corresponds to latent variables generated with a given structure, enforcing that the evening condition lies between day and night in the latent space, and the second row corresponds to a learned latent variable with no enforced structure. Given images at one appearance condition our method is able to smoothly transition the appearance to match a different weather and lighting condition, generating plausible intermediate visual appearances. Additional results for interpolation can be seen in the video on the project page: \href{https://ava-nvs.github.io}{https://ava-nvs.github.io}}
\label{fig_WeatherInterpolation}
\end{figure}

\begin{figure}[t]
\centering
\includegraphics[width=0.85\columnwidth]{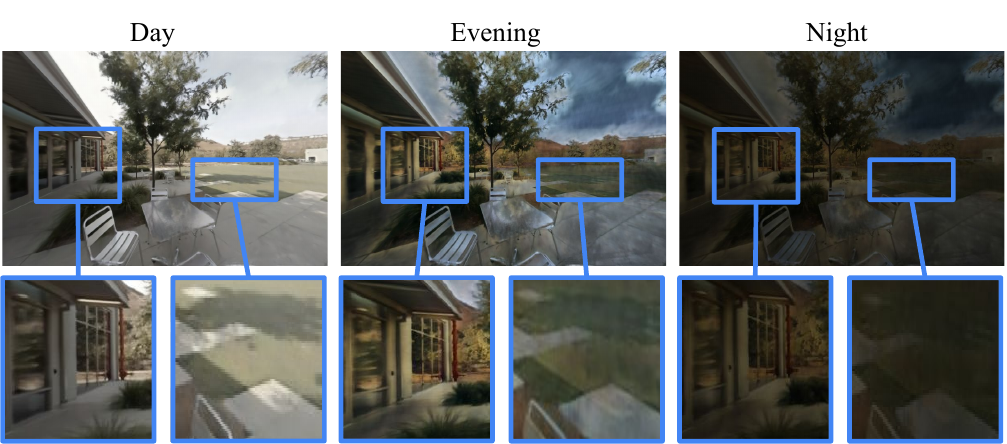}
\caption{Visual appearance change applied on a daytime scene from the Spaces dataset \cite{Spaces}. We observe that our method is able to make realistic appearance changes, such as adding sunlight on the background and light reflections in the windows for the evening condition and removing shadows for the night condition, without being trained on any scenes from this dataset.}
\label{fig_SpacesInterpolation}
\end{figure}

\paragraph{Qualitative Results.}
Our model is evaluated on the 38 evaluation scenes not seen during training.
The method is capable of synthesizing novel views using only a set of images with corresponding camera poses. Furthermore, it is able to adapt the visual appearance of the scene to specified weather and lighting conditions, without having access to observations of the scene under those target conditions. We show several qualitative examples of this. Fig.~\ref{fig_results_4seasons} shows that our method is able to change the visual appearance of images to match a target weather and lighting condition, and Fig.~\ref{fig_2dcomp} shows a comparison with other methods.

It also becomes possible to interpolate between two latent variables corresponding to different conditions by defining $\mathbf{z}_\alpha = \alpha \mathbf{z_c} + (1-\alpha) \mathbf{z_{c'}}$ for $\alpha \in [0,1]$. In Fig.~\ref{fig_WeatherInterpolation}, we observe that this enables getting realistic intermediate visual appearances that are not included in the original images. The model trained on appearance change of synthetic scenes can also be applied to change appearance of real scenes \cite{Spaces}, as seen in Fig.~\ref{fig_SpacesInterpolation} where we can see realistic appearance changes even though the model is not trained on any scenes from that dataset.

\footnotetext[1]{{We could not get satisfactory renderings for this condition, more details can be found in appendix \ref{sec:in2n_comp}.}}
\paragraph{Quantitative Results.}
We now show quantitative rendering quality results. We show PSNR, SSIM \cite{SSIM} and LPIPS \cite{LPIPS}, where the images with changed appearance are evaluated against the corresponding ground truth images for the target weather and lighting conditions. In Table~\ref{tab:results_2D_comp}, we show how our method performs on all possible combinations of source and target conditions. Using the same source and target conditions corresponds to novel view synthesis without appearance change, which, as expected, gives better metrics, but the gap is small for some combinations, e.g. comparing Day into Evening with Evening into Evening.
In Table~\ref{tab:Compare_w_2D}, we compare our method with several 2D style transfer methods. We see that our method outperforms the 2D methods on the performance metrics for all combinations. %, while giving competitive results for LPIPS.
We observe that performance varies for the different conditions and that adapting images from another condition into day is the most challenging, while transforming from day gives significantly higher performance for all methods. A comparison against Instruct-NeRF2NeRF \cite{in2n} was also performed, but results varied largely for different scenes and prompts. Further details are included in appendix \ref{sec:in2n_comp}. 

We show two consistency metrics \cite{chu2020learning} in Table~\ref{tab:Compare_w_2D_consistency}. If $(x_1, \hdots, x_n)$ and $(y_1, \hdots, y_n)$ are two image sequences rendered from the same pose sequences, we define $\text{tOF}=\| \text{OF}(y_{t+1}, y_t) - \text{OF}(x_{t+1}, x_t) \|_1$, where $\text{OF}$ is the optical flow computed via RAFT \cite{teed2020raft} and $\text{tLP}=\| \text{LPIPS}(y_{t+1}, y_t) - \text{LPIPS}(x_{t+1}, x_t) \|_1$. The metrics are low if the reference images and the rendered images yield similar optical flow and similar changes in $\text{LPIPS}$, which is assumed to correspond to a consistent rendering. We observe that our method significantly outperforms most of the 2D style transfer methods. 
Notably, Instruct-NeRF2NeRF exhibits poorer consistency results than anticipated, primarily stemming from two key factors. Firstly, the NeRF models generate low-quality novel views for some scenes. Secondly, there are inconsistent appearance changes in response to certain prompts, which results in unrealistic alterations that do not clearly preserve the scene content. CyEDA gives comparable consistency metrics for some scenarios, but gives less realistic rendered views as seen in Fig.~\ref{fig_2dcomp}
 and Table~\ref{tab:Compare_w_2D}.

\begin{table}[t]
\caption{Ablation Study comparing two approaches for generating latent appearance variables, comparing the similarity of rendered views with ground truth images (\textbf{\small{PSNR $\uparrow$ $\vert$ SSIM$\uparrow$ $\vert$ LPIPS$\downarrow$}}). We observe that both approaches give similar performance for changing appearance from one condition to another. }
\centering
\begin{adjustbox}{width=\linewidth,center}
\begin{tabular}{lcllcllcllcll}
%\hline
\cmidrule[2pt]{1-13}
\multicolumn{1}{c}{\multirow{2}{*}{Latent Variables}} & \multicolumn{12}{c}{Scenarios}                                                                                                                                	\\ \cline{2-13}
\multicolumn{1}{c}{}                              	& \multicolumn{3}{c}{Day to Night}            	& \multicolumn{3}{c}{Day to Evening}            	& \multicolumn{3}{c}{Day to Rain}            	& \multicolumn{3}{c}{Night to Day}            	\\ \cmidrule[2pt]{1-13} %\hline
Enforced structure                                	& \multicolumn{3}{c}{21.0 \hspace{-0.65em} \textcolor{gray}{$\vert$}\hspace{-0.27em} 0.56 \hspace{-0.65em} \textcolor{gray}{$\vert$}\hspace{-0.27em} 0.55} & \multicolumn{3}{c}{24.1 \hspace{-0.65em} \textcolor{gray}{$\vert$}\hspace{-0.27em} 0.75 \hspace{-0.65em} \textcolor{gray}{$\vert$}\hspace{-0.27em} 0.58} & \multicolumn{3}{c}{23.4 \hspace{-0.65em} \textcolor{gray}{$\vert$}\hspace{-0.27em} 0.71 \hspace{-0.65em} \textcolor{gray}{$\vert$}\hspace{-0.27em} 0.58} & \multicolumn{3}{c}{15.3 \hspace{-0.65em} \textcolor{gray}{$\vert$}\hspace{-0.27em} 0.56 \hspace{-0.65em} \textcolor{gray}{$\vert$}\hspace{-0.27em} 0.62} \\
No enforced structure                             	& \multicolumn{3}{c}{21.6 \hspace{-0.65em} \textcolor{gray}{$\vert$}\hspace{-0.27em} 0.57 \hspace{-0.65em} \textcolor{gray}{$\vert$}\hspace{-0.27em} 0.55} & \multicolumn{3}{c}{23.2 \hspace{-0.65em} \textcolor{gray}{$\vert$}\hspace{-0.27em} 0.71 \hspace{-0.65em} \textcolor{gray}{$\vert$}\hspace{-0.27em} 0.57} & \multicolumn{3}{c}{22.8 \hspace{-0.65em} \textcolor{gray}{$\vert$}\hspace{-0.27em} 0.70 \hspace{-0.65em} \textcolor{gray}{$\vert$}\hspace{-0.27em} 0.55} & \multicolumn{3}{c}{15.3 \hspace{-0.65em} \textcolor{gray}{$\vert$}\hspace{-0.27em} 0.56 \hspace{-0.65em} \textcolor{gray}{$\vert$}\hspace{-0.27em} 0.61} \\ \cmidrule[2pt]{1-13} %\hline
\end{tabular}
\end{adjustbox}
\label{tab:ablation}
\end{table}

\paragraph{Ablation Study.}
We compare two different ways of learning latent appearance variables $\mathbf{z}_{c} \in \mathbb{R}^d$. One approach is to initialize a random $d$-dimensional vector with no enforced structure for each condition as a learnable parameter that is optimized jointly with the rest of the model.
Another approach is to enforce structure by representing each condition as a fixed 2D-coordinate, placing them such that the evening condition is in between day and night, based on the assumption that one should pass through evening when going from day to night. 
These fixed 2D coordinates are then fed through a small learned fully-connected network to generate $\mathbf{z}_{c}$. Comparing the performance metrics for these two approaches, as can be seen in Table~\ref{tab:ablation}, shows that both approaches give similar performance. However, enforcing a structure on the latent space leads to more realistic lighting effects when interpolating, as can be seen in Fig. \ref{fig_WeatherInterpolation}, giving the appearance of a sunset. Based on this, it was decided to use the latent appearance variable with the enforced structure for our experiments.
The choice of dimension $d = 136$, for the latent appearance variable, was made by observing that a higher dimension leads to a better ability to handle local appearance changes, such as turning on street lamps and removing shadows. More details and qualitative examples can be found in appendix \ref{sec:GenLatent}.

\section{Conclusions}
We present a transformer based generalizable novel view synthesis method that allows for change of visual appearance without any scene-specific training. This is achieved by introducing a latent appearance variable that is used to change the visual appearance to match a given weather and lighting condition while keeping the scene structure unchanged.  We also introduce a synthetic dataset based on CARLA for training and evaluating the methods and present experiments that show that this method is able to change the visual appearance of both synthetic and real scenes, to match a specified weather and lighting condition without any scene-specific training. The generated latent variables also make it possible to smoothly interpolate between different weather and lighting conditions. Compared to 2D style transfer, our method is view consistent by design. We experimentally show that our method outperforms multiple 2D style transfer methods, both in terms of rendering quality and that the rendering of nearby views are more consistent. A comparison with Instruct-NeRF2NeRF shows that our method is more robust in providing desired appearance changes while ensuring multi-view consistency and preserving scene content. Our generalizable approach is also more flexible, not requiring training a NeRF model for each scene, and also allows using fewer input images.

\subsubsection{Acknowledgements}
This work received full support from the Wallenberg AI, Autonomous Systems, and Software Program (WASP) funded by the Knut and Alice Wallenberg Foundation. Computational resources were provided by the National Academic Infrastructure for Supercomputing in Sweden (NAISS) at Chalmers Centre for Computational Science and Engineering (C3SE), partially funded by the Swedish Research Council under grant agreement no. 2022-06725.

\bibliographystyle{splncs04}
\bibliography{egbib}

\begin{thebibliography}{10}
\providecommand{\url}[1]{\texttt{#1}}
\providecommand{\urlprefix}{URL }
\providecommand{\doi}[1]{https://doi.org/#1}

\bibitem{Mip-NeRF}
Barron, J.T., Mildenhall, B., Tancik, M., Hedman, P., Martin-Brualla, R., Srinivasan, P.P.: Mip-nerf: A multiscale representation for anti-aliasing neural radiance fields. ICCV  (2021)

\bibitem{beh2022cyeda}
Beh, J.C., Ng, K.W., Kew, J.L., Lin, C.T., Chan, C.S., Lai, S.H., Zach, C.: Cyeda: Cycle-object edge consistency domain adaptation. In: ICIP (2022)

\bibitem{brooks2022instructpix2pix}
Brooks, T., Holynski, A., Efros, A.A.: Instructpix2pix: Learning to follow image editing instructions. In: CVPR (2023)

\bibitem{brown2020language}
Brown, T., Mann, B., Ryder, N., Subbiah, M., Kaplan, J.D., Dhariwal, P., Neelakantan, A., Shyam, P., Sastry, G., Askell, A., et~al.: Language models are few-shot learners. NeurIPS  \textbf{33},  1877--1901 (2020)

\bibitem{MVSNeRF}
Chen, A., Xu, Z., Zhao, F., Zhang, X., Xiang, F., Yu, J., Su, H.: Mvsnerf: Fast generalizable radiance field reconstruction from multi-view stereo. In: ICCV (2021)

\bibitem{chen2020reusing}
Chen, R., Huang, W., Huang, B., Sun, F., Fang, B.: Reusing discriminators for encoding: Towards unsupervised image-to-image translation. In: CVPR (2020)

\bibitem{chu2020learning}
Chu, M., Xie, Y., Mayer, J., Leal-Taix{\'e}, L., Thuerey, N.: Learning temporal coherence via self-supervision for gan-based video generation. ACM Transactions on Graphics (TOG)  \textbf{39}(4),  75--1 (2020)

\bibitem{dhariwal2021diffusion}
Dhariwal, P., Nichol, A.: Diffusion models beat gans on image synthesis. NeurIPS  (2021)

\bibitem{VIT}
Dosovitskiy, A., Beyer, L., Kolesnikov, A., Weissenborn, D., Zhai, X., Unterthiner, T., Dehghani, M., Minderer, M., Heigold, G., Gelly, S., Uszkoreit, J., Houlsby, N.: An {Image} is {Worth} 16x16 {Words}: {Transformers} for {Image} {Recognition} at {Scale}. In: ICLR (2022)

\bibitem{CARLA}
Dosovitskiy, A., Ros, G., Codevilla, F., Lopez, A., Koltun, V.: {CARLA}: {An} open urban driving simulator. In: Proceedings of the 1st Annual Conference on Robot Learning. pp. 1--16 (2017)

\bibitem{du2023LearningRenderNovel}
Du, Y., Smith, C., Tewari, A., Sitzmann, V.: Learning {To} {Render} {Novel} {Views} {From} {Wide}-{Baseline} {Stereo} {Pairs}. In: CVPR (2023)

\bibitem{Spaces}
Flynn, J., Broxton, M., Debevec, P., DuVall, M., Fyffe, G., Overbeck, R., Snavely, N., Tucker, R.: Deepview: View synthesis with learned gradient descent. In: CVPR (2019)

\bibitem{goodfellow2014generative}
Goodfellow, I., Pouget-Abadie, J., Mirza, M., Xu, B., Warde-Farley, D., Ozair, S., Courville, A., Bengio, Y.: Generative adversarial networks. Communications of the ACM  \textbf{63}(11),  139--144 (2020)

\bibitem{styleNeRF}
Gu, J., Liu, L., Wang, P., Theobalt, C.: Stylene{RF}: A style-based 3d aware generator for high-resolution image synthesis. In: ICLR (2022)

\bibitem{in2n}
Haque, A., Tancik, M., Efros, A., Holynski, A., Kanazawa, A.: Instruct-nerf2nerf: Editing 3d scenes with instructions. In: ICCV (2023)

\bibitem{hertz2022prompt}
Hertz, A., Mokady, R., Tenenbaum, J., Aberman, K., Pritch, Y., Cohen-Or, D.: Prompt-to-prompt image editing with cross attention control. arXiv preprint arXiv:2208.01626  (2022)

\bibitem{ho2020denoising}
Ho, J., Jain, A., Abbeel, P.: Denoising diffusion probabilistic models. neuriPS  (2020)

\bibitem{ArbitraryStyleTransfer}
Huang, X., Belongie, S.: Arbitrary style transfer in real-time with adaptive instance normalization. In: ICCV (Oct 2017)

\bibitem{StylizedNeRF}
Huang, Y.H., He, Y., Yuan, Y.J., Lai, Y.K., Gao, L.: Stylizednerf: Consistent 3d scene stylization as stylized nerf via 2d-3d mutual learning. In: CVPR (2022)

\bibitem{pix2pix2017}
Isola, P., Zhu, J.Y., Zhou, T., Efros, A.A.: Image-to-image translation with conditional adversarial networks. In: CVPR (2017)

\bibitem{AdaNeRF}
Kurz, A., Neff, T., Lv, Z., Zollh\"{o}fer, M., Steinberger, M.: Adanerf: Adaptive sampling for real-time rendering of neural radiance fields. In: ECCV (2022)

\bibitem{kwon2022diffusion}
Kwon, G., Ye, J.C.: Diffusion-based image translation using disentangled style and content representation. arXiv preprint arXiv:2209.15264  (2022)

\bibitem{Neuray}
Liu, Y., Peng, S., Liu, L., Wang, Q., Wang, P., Christian, T., Zhou, X., Wang, W.: Neural rays for occlusion-aware image-based rendering. In: CVPR (2022)

\bibitem{NeRF-W}
Martin-Brualla, R., Radwan, N., Sajjadi, M.S.M., Barron, J.T., Dosovitskiy, A., Duckworth, D.: Nerf in the wild: Neural radiance fields for unconstrained photo collections. In: CVPR (2021)

\bibitem{NeRF}
Mildenhall, B., Srinivasan, P.P., Tancik, M., Barron, J.T., Ramamoorthi, R., Ng, R.: Nerf: Representing scenes as neural radiance fields for view synthesis. In: ECCV (2020)

\bibitem{InstantNeuralGraphics}
M\"uller, T., Evans, A., Schied, C., Keller, A.: Instant neural graphics primitives with a multiresolution hash encoding. ACM Trans. Graph.  \textbf{41}(4),  102:1--102:15 (Jul 2022)

\bibitem{RegNeRF}
Niemeyer, M., Barron, J.T., Mildenhall, B., Sajjadi, M.S.M., Geiger, A., Radwan, N.: Regnerf: Regularizing neural radiance fields for view synthesis from sparse inputs. In: CVPR (2022)

\bibitem{park2019arbitrary}
Park, D.Y., Lee, K.H.: Arbitrary style transfer with style-attentional networks. In: CVPR (2019)

\bibitem{CLIP}
Radford, A., Kim, J.W., Hallacy, C., Ramesh, A., Goh, G., Agarwal, S., Sastry, G., Askell, A., Mishkin, P., Clark, J., Krueger, G., Sutskever, I.: Learning transferable visual models from natural language supervision. In: ICML (2021)

\bibitem{rombach2022high}
Rombach, R., Blattmann, A., Lorenz, D., Esser, P., Ommer, B.: High-resolution image synthesis with latent diffusion models. In: CVPR (2022)

\bibitem{U-Net}
Ronneberger, O., P.Fischer, Brox, T.: U-net: Convolutional networks for biomedical image segmentation. In: Medical Image Computing and Computer-Assisted Intervention (MICCAI). LNCS, vol.~9351, pp. 234--241. Springer (2015)

\bibitem{saharia2022palette}
Saharia, C., Chan, W., Chang, H., Lee, C., Ho, J., Salimans, T., Fleet, D., Norouzi, M.: Palette: Image-to-image diffusion models. In: ACM SIGGRAPH (2022)

\bibitem{SRN}
Sajjadi, M.S., Meyer, H., Pot, E., Bergmann, U., Greff, K., Radwan, N., Vora, S., Lucic, M., Duckworth, D., Dosovitskiy, A., Uszkoreit, J., Funkhouser, T., Tagliasacchi, A.: Scene {Representation} {Transformer}: {Geometry}-{Free} {Novel} {View} {Synthesis} {Through} {Set}-{Latent} {Scene} {Representations}. In: CVPR (2022)

\bibitem{GPNR}
Suhail, M., Esteves, C., Sigal, L., Makadia, A.: Generalizable {Patch}-{Based} {Neural} {Rendering}. In: ECCV (2022)

\bibitem{GNT}
T, M.V., Wang, P., Chen, X., Chen, T., Venugopalan, S., Wang, Z.: Is attention all that ne{RF} needs? In: ICLR (2023)

\bibitem{Block-NeRF}
Tancik, M., Casser, V., Yan, X., Pradhan, S., Mildenhall, B., Srinivasan, P.P., Barron, J.T., Kretzschmar, H.: Block-nerf: Scalable large scene neural view synthesis. In: CVPR (2022)

\bibitem{Nerfstudio}
Tancik, M., Weber, E., Ng, E., Li, R., Yi, B., Wang, T., Kristoffersen, A., Austin, J., Salahi, K., Ahuja, A., Mcallister, D., Kerr, J., Kanazawa, A.: Nerfstudio: A modular framework for neural radiance field development. In: ACM SIGGRAPH 2023. Association for Computing Machinery, New York, NY, USA

\bibitem{teed2020raft}
Teed, Z., Deng, J.: Raft: Recurrent all-pairs field transforms for optical flow. In: ECCV (2020)

\bibitem{transformer}
Vaswani, A., Shazeer, N., Parmar, N., Uszkoreit, J., Jones, L., Gomez, A.N., Kaiser, Å., Polosukhin, I.: Attention is {All} you {Need}. In: NIPS (2017)

\bibitem{CLIP-NeRF}
Wang, C., Chai, M., He, M., Chen, D., Liao, J.: Clip-nerf: Text-and-image driven manipulation of neural radiance fields. In: CVPR (2022)

\bibitem{NeRF-Art}
Wang, C., Jiang, R., Chai, M., He, M., Chen, D., Liao, J.: Nerf-art: Text-driven neural radiance fields stylization. IEEE Transactions on Visualization and Computer Graphics pp. 1--15 (2023)

\bibitem{TransNeRF}
Wang, D., Cui, X., Salcudean, S., Wang, Z.J.: Generalizable neural radiance fields for novel view synthesis with transformer (2022)

\bibitem{IBRNet}
Wang, Q., Wang, Z., Genova, K., Srinivasan, P.P., Zhou, H., Barron, J.T., Martin-Brualla, R., Snavely, N., Funkhouser, T.: Ibrnet: Learning multi-view image-based rendering. In: CVPR (2021)

\bibitem{wang2018high}
Wang, T.C., Liu, M.Y., Zhu, J.Y., Tao, A., Kautz, J., Catanzaro, B.: High-resolution image synthesis and semantic manipulation with conditional gans. In: CVPR (2018)

\bibitem{SSIM}
Wang, Z., Bovik, A., Sheikh, H., Simoncelli, E.: Image quality assessment: from error visibility to structural similarity. IEEE Transactions on Image Processing  \textbf{13}(4),  600--612 (2004)

\bibitem{pixelNeRF}
Yu, A., Ye, V., Tancik, M., Kanazawa, A.: pixelnerf: Neural radiance fields from one or few images. In: CVPR (2021)

\bibitem{ARF}
Zhang, K., Kolkin, N., Bi, S., Luan, F., Xu, Z., Shechtman, E., Snavely, N.: Arf: Artistic radiance fields (2022)

\bibitem{LPIPS}
Zhang, R., Isola, P., Efros, A.A., Shechtman, E., Wang, O.: The unreasonable effectiveness of deep features as a perceptual metric. In: CVPR (2018)

\bibitem{CycleGAN2017}
Zhu, J.Y., Park, T., Isola, P., Efros, A.A.: Unpaired image-to-image translation using cycle-consistent adversarial networks. In: ICCV (2017)

\bibitem{zhu2017toward}
Zhu, J.Y., Zhang, R., Pathak, D., Darrell, T., Efros, A.A., Wang, O., Shechtman, E.: Toward multimodal image-to-image translation. NIPS  (2017)

\end{thebibliography}

\newpage
\appendix
\section*{Appendix}

\noindent A supplementary video is available at \href{https://ava-nvs.github.io}{https://ava-nvs.github.io}. The video results are discussed in \ref{sec:video}.  In \ref{sec:GenLatent} we give more detail regarding the generation of latent appearance variables. In \ref{sec:in2n_comp} we give additional details regarding the comparison with Instruct-NeRF2NeRF\cite{in2n}.  In \ref{sec:MoreScenarios} we show qualitative comparisons for additional scenarios.

\section{Video}
\label{sec:video}
The supplementary video presents additional findings regarding interpolation, demonstrating our method's capacity to given images at one appearance condition smoothly transition the appearance to match a different weather and lighting condition while ensuring multi-view consistency. 

Furthermore, the video includes a temporal consistency comparison with applying 2D style transfer on rendered images, for the Day to Night scenario. Our observations reveal that nearly all 2D methods exhibit some degree of flickering and temporal inconsistencies, with DiffuseIT~\cite{kwon2022diffusion} displaying the most significant issues, frequently hallucinating objects or structures. CyEDA~\cite{beh2022cyeda} appears to generate random unexpected bright spots. Of the 2D methods only Palette~\cite{saharia2022palette} successfully learns to fully activate street lamps, but noticeable pixel intensity fluctuations result in inconsistent renderings. In contrast, our method delivers both multi-view consistency and realistic lighting changes.

Finally the video also contains a comparison with the Instruct-NeRF2NeRF method for the Day to Night and Day to Rain scenarios. We can observe that Instruct-NeRF2NeRF method gives multiview consistent renderings, but it struggles with increased blurriness and cloudy artifacts. We can for the Day to Night scenario observe that the method struggles with clearly preserving the scene content. More details about the Instruct-NeRF2NeRF comparison can be found in the next section, including results for additional prompts.

\section{Generating Latent Variables}\label{sec:GenLatent}
We propose two approaches for generating latent appearance variables $\mathbf{z_{c}} \in \mathbb{R}^d$. One approach is to initialize a random $d$-dimensional vector for each condition and then include it as a learnable parameter that is optimized jointly with the rest of the model. In this case, the latent variables are fully learned with no enforced structure.  

\begin{figure}[tbh]
\centering
\includegraphics[width=0.7\columnwidth]{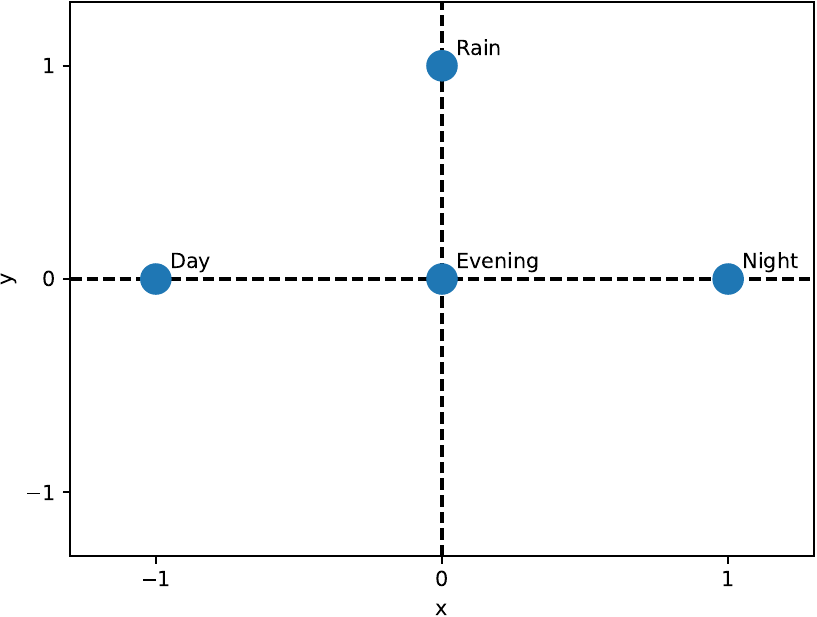}
\caption{Chosen fixed 2D coordinates for each condition, ensuring that one passes through the evening condition when interpolating between the day and night conditions. Rain is placed on a separate axis since it corresponds to appearance change not directly connected to variations in daylight.}
\label{fig_2D_Coords}
\end{figure}

\begin{figure}[tb]
\centering
\includegraphics[width=0.8\columnwidth]{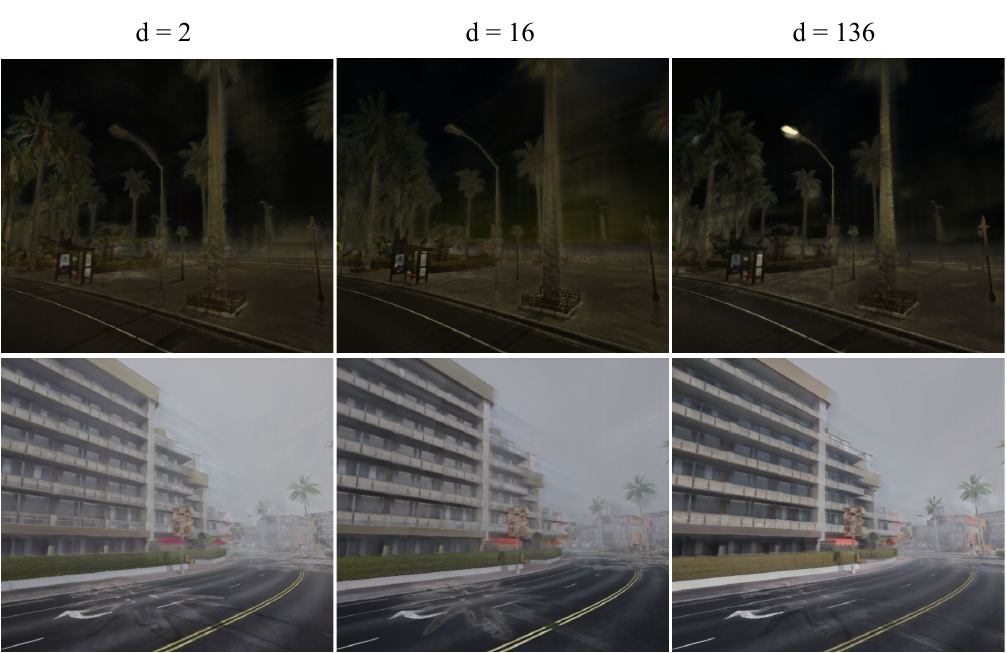}
\caption{Qualitative comparison of rendered views with changed appearance for different sizes $d$ of the latent appearance variable $\mathbf{z_{c}}$. We observe that a higher value of $d$ leads to better local appearance changes in rendered views, such as turning on street lamps and removing shadows.}
\label{fig_d_comp}
\end{figure}

Another approach is to enforce structure on the latent appearance variables by defining fixed 2D-coordinates $c$ corresponding to each condition that is then passed through a small fully-connected network to generate $\mathbf{z_{c}} = f_z(c)$, where the parameters of this additional fully-connected network are learned jointly with the rest of the model. For our case with four weather and lighting conditions, we define the fixed 2D coordinates as shown in Fig.~\ref{fig_2D_Coords}. 
% \begin{equation}
% c_{day} = \begin{bmatrix}
%  -1\\ 0
% \end{bmatrix}, \;  c_{evening} = \begin{bmatrix}
%  0\\ 0
% \end{bmatrix},
%  \;c_{rain} = \begin{bmatrix}
%  0\\  1
% \end{bmatrix},\; c_{night} = \begin{bmatrix}
%  1\\ 0
% \end{bmatrix},
% \end{equation}
The reason behind this placement is to get the desired behavior when interpolating between two conditions, ensuring that the evening condition is passed through when interpolating between day and night conditions, and places rain on a separate axis since it corresponds to appearance change not directly connected to variations in daylight. The fully connected network $f_z(c)$ takes in a 2D coordinate corresponding to a condition and outputs a latent appearance variable $\mathbf{z_{c}}$ of dimension $d$. For the performed experiments, we used $d = 136$, and two hidden layers of size $16$ and $68$, respectively. The choice of $d$ was made after testing different values and observing that a higher dimension leads to better ability to handle local appearance changes such as turning on lamps and removing shadows, as seen in Fig.\ \ref{fig_d_comp}.

Comparing performance metrics for learnable latent variables with no enforced structure in Table~\ref{tab:results_no_structure} with the ones in Table~\ref{tab:results_2D_comp} where latent appearance variables with enforced structure are used, shows that both approaches for generating the latent variable $\mathbf{z_{c}}$ give similar performance when changing appearance from one condition to another. However, enforcing a structure on the latent space leads to more realistic lighting effects when interpolating between two conditions, as can be seen in Fig.~\ref{fig_WeatherInterpolation}, giving the appearance of a sunset. Based on this, it was decided to use the latent appearance variable with the enforced structure for our experiments.
% However, enforcing a structure on the latent space leads to more realistic lighting effects when interpolating, as can be seen in Fig.\ 5 in the main paper. Based on this it was decided to use the latent appearance variable with enforced structure for our experiments.

\begin{table}[t]
\caption{Comparison of similarity of rendered views when using learnable latent variables with no enforced structure. Comparing with ground truth images for all combinations of weather and lighting conditions (\textbf{\small{PSNR$\uparrow$ $\vert$ SSIM$\uparrow$ $\vert$ LPIPS$\downarrow$}}). The values along the diagonal correspond to novel view synthesis without appearance change.}
\centering
%\small
\begin{adjustbox}{width=\linewidth,center}
\begin{tabular}{lcccc}
\cmidrule[2pt]{1-5}
%\hline
         	& From Day                                                                                             	& From Night                                                                                           	& From Evening                                                                                         	& From Rain                                                                                            	\\ \cmidrule[2pt]{1-5} %\hline
Into Day 	& 23.3 \textcolor{gray}{$\vert$} 0.76 \textcolor{gray}{$\vert$} 0.61 & 15.3 \textcolor{gray}{$\vert$} 0.56 \textcolor{gray}{$\vert$} 0.61 & 16.4 \textcolor{gray}{$\vert$} 0.61 \textcolor{gray}{$\vert$} 0.60 & 15.9 \textcolor{gray}{$\vert$} 0.61 \textcolor{gray}{$\vert$} 0.61 \\
Into Night   & 21.6 \textcolor{gray}{$\vert$} 0.57 \textcolor{gray}{$\vert$} 0.55 & 28.1 \textcolor{gray}{$\vert$} 0.73 \textcolor{gray}{$\vert$} 0.55 &  21.1 \textcolor{gray}{$\vert$} 0.56 \textcolor{gray}{$\vert$} 0.54 & 21.4 \textcolor{gray}{$\vert$} 0.58 \textcolor{gray}{$\vert$} 0.55 \\
Into Evening & 23.2 \textcolor{gray}{$\vert$} 0.71 \textcolor{gray}{$\vert$} 0.57 & 19.9 \textcolor{gray}{$\vert$} 0.47 \textcolor{gray}{$\vert$} 0.58 & 23.1 \textcolor{gray}{$\vert$} 0.55 \textcolor{gray}{$\vert$} 0.57 & 20.3 \textcolor{gray}{$\vert$} 0.50 \textcolor{gray}{$\vert$} 0.57 \\
Into Rain	& 22.8 \textcolor{gray}{$\vert$} 0.70 \textcolor{gray}{$\vert$} 0.55 & 20.9 \textcolor{gray}{$\vert$} 0.66 \textcolor{gray}{$\vert$} 0.56  & 21.3 \textcolor{gray}{$\vert$} 0.69 \textcolor{gray}{$\vert$} 0.57  & 23.1 \textcolor{gray}{$\vert$} 0.55 \textcolor{gray}{$\vert$} 0.58 \\ \cmidrule[2pt]{1-5} %\hline
\end{tabular}
\end{adjustbox}
\label{tab:results_no_structure}
\end{table}

\section{Comparison with Instruct-NeRF2NeRF}
\label{sec:in2n_comp}
\begin{figure}[tbh]
\centering
\includegraphics[width=1\columnwidth]{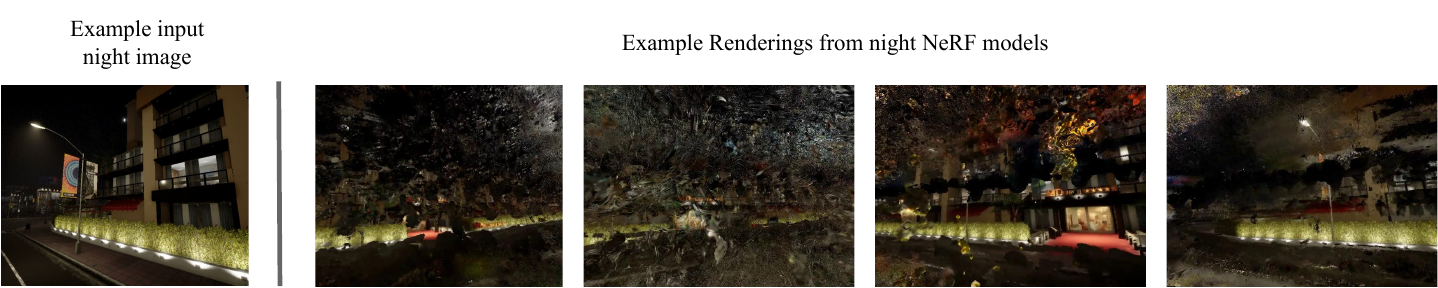}
\caption{Renderings from Night time NeRF scenes were of very low quality, so consistency metrics were therefore not computed for the Night to Day condition.}
\label{fig_night_renderings}
\end{figure}

We used the official Instruct-NeRF2NeRF implementation. This implementation uses the Nerfstudio \cite{Nerfstudio} Nerfacto NeRF model. The quality of the Nerfacto models when using 10 images was very poor, so we increased the number of images in the sequence to 25. There were still issues of the quality of the rendered views for some scenes, especially for the night scenes, as can be seen in Fig.~\ref{fig_night_renderings}. This is the reason why consistency metrics for the Night to Day scenario for the Instruct-NeRF2NeRF method are not included in Table~\ref{tab:Compare_w_2D_consistency}. 
%Mention which prompts have been used. Show examples for some more scenes. Mention that does not work for some scenes for which instruct-pix2pix does work.

\begin{figure}[tbh]
\centering
\includegraphics[width=1\columnwidth]{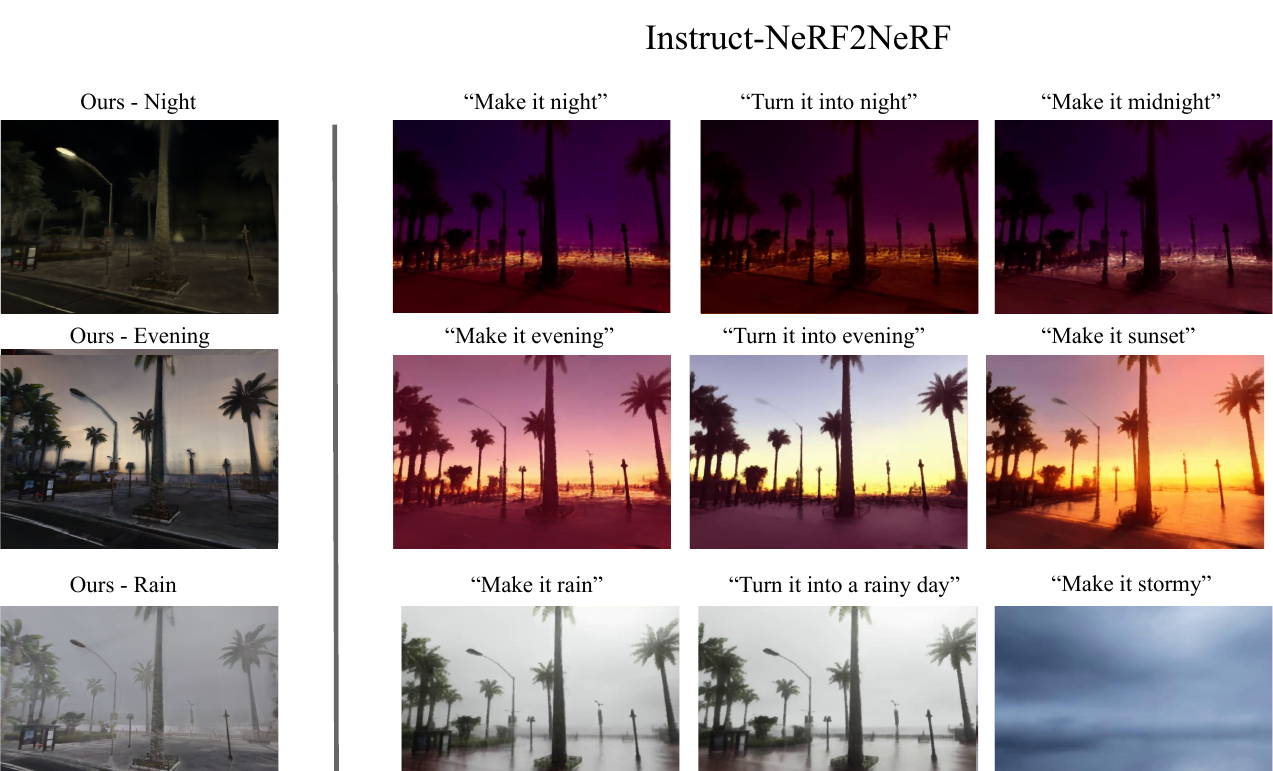}
\caption{Comparing Instruct-NeRF2NeRF for different prompts.}
\label{fig_scene6_prompts}
\end{figure}

\begin{figure}[tbh]
\centering
\includegraphics[width=1\columnwidth]{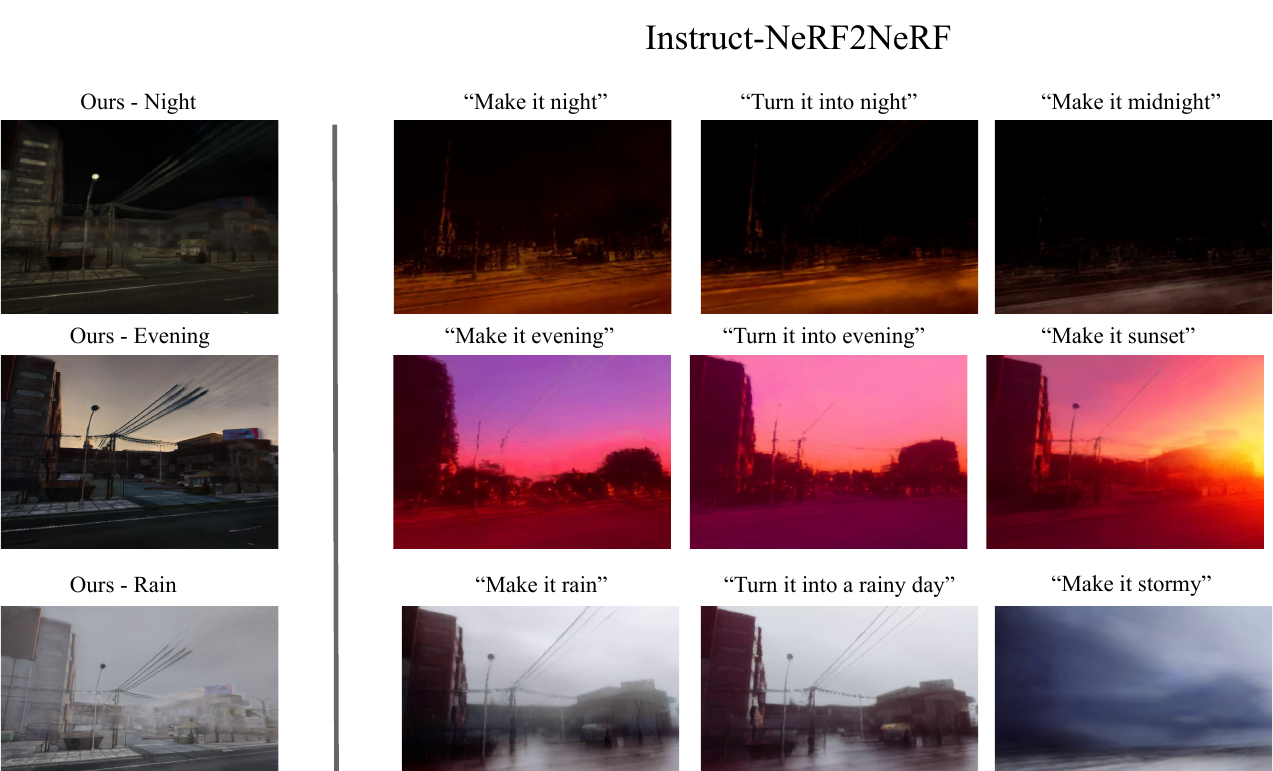}
\caption{Comparing Instruct-NeRF2NeRF for different prompts.}
\label{fig_scene73_prompts}
\end{figure}

Figures ~\ref{fig_scene6_prompts} and ~\ref{fig_scene73_prompts} show results for Instruct-NeRF2NeRF for additional prompts. This shows that the quality of rendered views can vary strongly based on the prompt that is used, and some prompts such as "Make it midnight" and "Make it stormy" led to the scene content almost completely disappearing. The figures also show that the same prompt can result in differing appearance changes when used on different scenes, e.g. the prompt "Turn it into evening" leads to images with very different color schemes for the two different scenes. In contrast our method gives similar types of appearance change for different scenes.

\section{Comparisons for additional scenarios}
\label{sec:MoreScenarios}

\begin{figure*}[!htb]
\centering

\includegraphics[width=\textwidth]{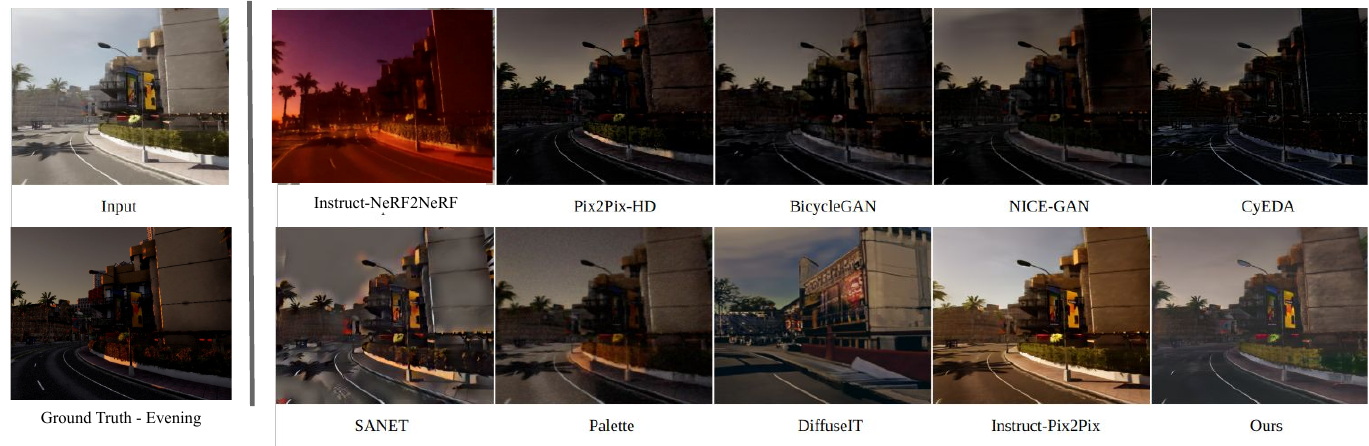}

\caption{Comparison with other methods for Day to Evening scenario.}
% Our method is able to change the visual appearance appropriately, in part by removing shadows. Other than our method, the shadows of trees and street lamps in the images generated by most of the 2D methods are still clearly noticeable, but they should be less visible in the evening when there is no direct sunlight. Pix2Pix-HD~\cite{wang2018high} seems to be close to our result, but our model can achieve the best quantitative results as seen in Table~1 in the main paper. DiffuseIT~\cite{kwon2022diffusion} can provide visually plausible results for individual images, but there are hallucinations that do not exist in the original images.
% \vspace*{0in}
\label{fig_2dcomp_D2E}
\end{figure*}

\begin{figure*}[!htb]
% \centering
% \includegraphics[width=\textwidth]{figures/result_comparison/D2R_comparison_supcaption_all_top.png}
% \includegraphics[width=\textwidth]{figures/result_comparison/D2R_comparison_supcaption_all_btm.png}
\includegraphics[width=\textwidth]{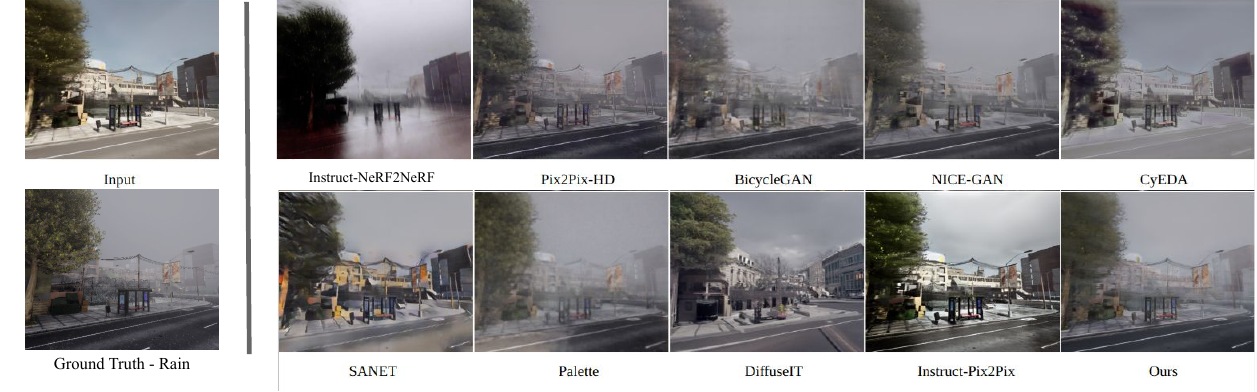}

\caption{Comparison with other methods for Day to Rain scenario.}
% \caption{Comparing our method with applying 2D style transfer on rendered images for the Day to Rain scenario. }
% Our method is able to change the visual appearance appropriately while clearly preserving scene content. In this scenario, all the 2D methods give more realistic renderings, except DiffuseIT~\cite{kwon2022diffusion}, which hallucinates content that does not exist in the original images.
% \vspace*{3in}
\label{fig_2dcomp_D2R}
\end{figure*}

\begin{figure*}[!t]
\centering
\includegraphics[width=\textwidth]{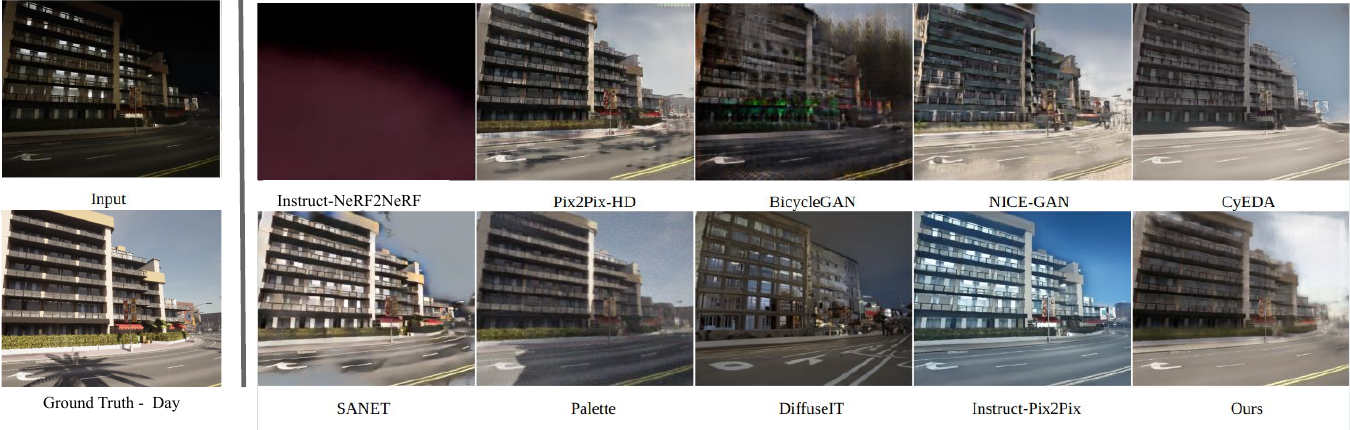}

\caption{Comparison with other methods for Night to Day scenario. We observe that the main reason for the Instruct-NeRF2NeRF methods low quality for this scenario is lacking quality in the Nerfacto \cite{Nerfstudio} NeRF models trained on night images.}

% \caption{Comparing our method with applying 2D style transfer on rendered images for the difficult Night to Day scenario.}
% Our method is able to change the visual appearance appropriately, in part by deactivating interior lighting inside the buildings. Most of the 2D methods struggle to provide structure-consistent results, and for some, the sky is not even bright. Instruct-Pix2Pix\cite{brooks2022instructpix2pix} achieves plausible results at first sight, but the interior lighting is still shining clearly, which should not be the case at day time.
\vspace*{3in}
\label{fig_2dcomp_N2D}
\end{figure*}

We further compare against other methods for additional scenarios, including Day to Evening (Fig.~\ref{fig_2dcomp_D2E}), Day to Rain (Fig.~\ref{fig_2dcomp_D2R}), and the most challenging scenario Night to Day (Fig.~\ref{fig_2dcomp_N2D}). Our method is able to clearly preserve content and 3D consistency while making appropriate adjustments to the appearance of the scene. For instance, we effectively eliminate shadows from images for the Day to Evening scenario. In the Day to Rain scenario, our model maintains scene content while altering visual appearance, ensuring multi-view consistency. Lastly, our model can even learn to deactivate interior lighting in buildings in the most challenging Night to Day scenario.

% [!htb]

\end{document}